\documentclass[letterpaper, 10 pt, conference]{ieeeconf}  

\IEEEoverridecommandlockouts                              

\usepackage{colortbl}
\usepackage[letterpaper, left=0.67in, right=0.67in, bottom=0.6in, top=0.79in]{geometry}
\usepackage{url}
\usepackage{graphicx}
\usepackage{amsmath}
\usepackage{caption}
\usepackage{subcaption}
\usepackage{amssymb}
\usepackage{multirow}
\usepackage{array}
\usepackage[linesnumbered,ruled,vlined]{algorithm2e}
\usepackage{booktabs}
\DeclareCaptionFont{CaptionFontSize}{\fontsize{8pt}{8pt}\selectfont}
\title{\LARGE \bf
Learn to Navigate Maplessly with Varied LiDAR Configurations: A Support Point-Based Approach
}

\author{Wei Zhang, Ning Liu, and Yunfeng Zhang 
	\thanks{Personal use of this material is permitted.  Permission from IEEE must be obtained for all other uses, in any current or future media, including reprinting/republishing this material for advertising or promotional purposes, creating new collective works, for resale or redistribution to servers or lists, or reuse of any copyrighted component of this work in other works.}%
}

\begin{document}

\maketitle
\thispagestyle{empty}
\pagestyle{empty}

\begin{abstract}

Deep reinforcement learning (DRL) demonstrates great potential in mapless navigation domain. However, such a navigation model is normally restricted to a fixed configuration of the range sensor because its input format is fixed. In this paper, we propose a DRL model that can address range data obtained from different range sensors with  different installation positions. Our model first extracts the goal-directed features from each obstacle point. Subsequently, it chooses global obstacle features from all point-feature candidates and uses these features for the final decision. As only a few points are used to support the final decision, we refer to these points as support points and our approach as support point-based navigation (SPN). Our model can handle data from different LiDAR setups and demonstrates good performance in simulation and real-world experiments. Moreover, it shows great potential in crowded scenarios with small obstacles when using a high-resolution LiDAR.

\end{abstract}

\section{INTRODUCTION}

Deep reinforcement learning (DRL) [1], which utilizes deep neural networks (DNN) to approximate navigation policy with reinforcement learning, has been widely applied in robot navigation domain. The DRL model shows great advantage in mapless navigation because it can directly handle the range data without any hand-engineered navigation rules compared to fuzzy reactive control [2]. It can also drive the robot out of small local-minimum areas [3] compared to artificial potential field approach [4]. Moreover, compared to the commonly used local planner in SLAM [5], such as dynamic window approach (DWA) [6], the DRL model requires neither a local map nor time-consuming forward simulation, and it can adjust its policy quickly when facing moving obstacles [7]. 

The most widely used sensors for DRL-based navigation are range sensors. With a range sensor, such as LiDAR, the model trained in simulation can be directly deployed to a real robot without any fine-tuning. Tai et al. [8] train their model in simulation scenarios, and the trained model can generate navigation policy using only ten laser beams. Xie et al. [9] introduce a PID controller to accelerate the training of DRL in simulation. This model is built with convolutional neural networks (CNN), which takes 512 laser beams as input. Pfeiffer et al. [10] use min-pooling to down-sample 36 distance values from 1080 laser beams and adopt a fully-connected network structure to enhance the generalization performance of their model. The min-pooling operation can help handle more laser points and enhance safety during navigation. However, it sacrifices much local information and may lead to a suboptimal policy [10]. Lim et al. [11] apply Gaussian process regression to predict the reward functions in DRL from expert-demonstrated data. Although this approach is data-efficient, the trained model only takes ten measurements from a 360-degree LiDAR for reducing the input space, which may cause the robot to collide with small obstacles.

\begin{figure}[tb]
	\centering
	\begin{subfigure}[h]{.24\linewidth}
		\centering
		\centerline{\includegraphics[width=0.98\linewidth]{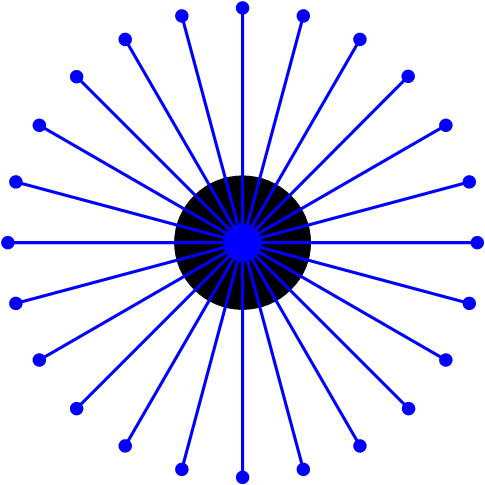}}  
		\caption{}
		\label{fig:1a}
	\end{subfigure}
	\hfill
	\begin{subfigure}[h]{.24\linewidth}
		\centering
		\centerline{\includegraphics[width=0.98\linewidth]{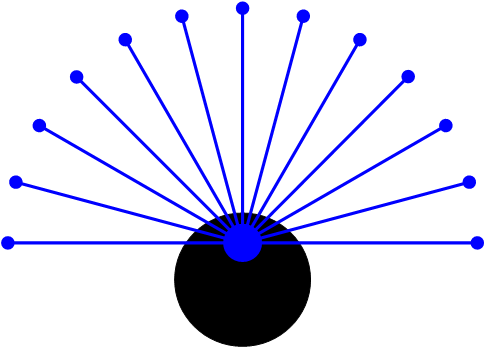}}
		\caption{}
		\label{fig:1b}
	\end{subfigure}
	\hfill
	\begin{subfigure}[h]{.24\linewidth}
		\centering
		\includegraphics[width=0.98\linewidth]{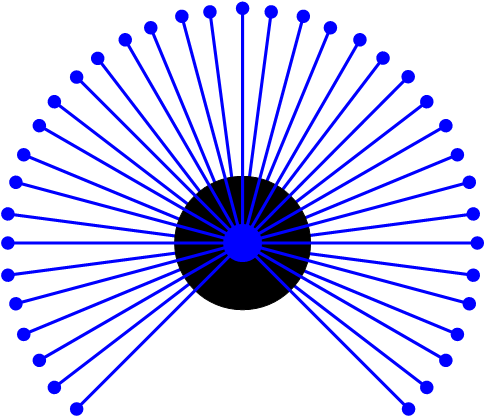}  
		\caption{}
		\label{fig:1c}
	\end{subfigure}
	\hfill
	\begin{subfigure}[h]{.24\linewidth}
		\centering
		\includegraphics[width=0.98\linewidth]{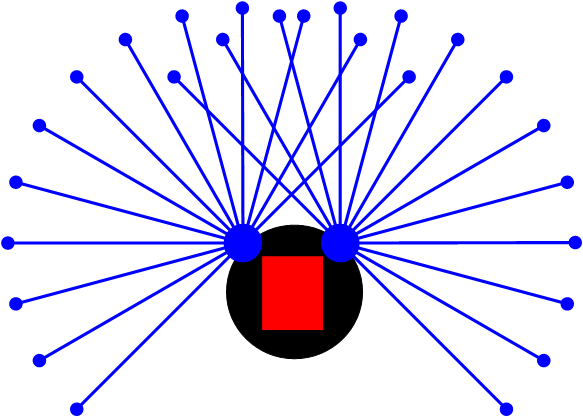}  
		\caption{}
		\label{fig:1d}
	\end{subfigure}
	\caption{Examples of changing the configuration of LiDAR used for training. (a) the LiDAR used for training, (b) LiDAR with a different FOV and position, (c) LiDAR with a different FOV and angular resolution, (d) two LiDARs are mounted on two sides to allow the robot carry items (the red block).}
	\label{fig:1}
\end{figure}

Although these approaches work well in robot navigation applications, the trained DNN model requires fixed parameters of the employed sensor, i.e., sensor position, field of view (FOV), and angular resolution. These parameters may change due to different task requirements. Fig. 1 shows examples of potential parameter changes of the LiDAR sensor. Once these parameters change, the original model cannot directly handle the new sensor readings. The unknown values in the 1D input need to be padded. However, it is difficult to ensure the accuracy of the padded value, especially when the new sensor has a smaller FOV and low angular resolution. Building a local grid map and inputting the map into a CNN model can allow the change of LiDAR configurations. However, map building requires additional computation, and the dimension of a high-resolution map is huge, which requires a high-performance computer for processing the data with CNN in real-world applications [12]. Besides, training a network with high-dimensional and sparse inputs is much harder than the low-dimensional dense inputs. Existing map-based models only operate on low-resolution maps ([12], [13] and [14]), which are not precise enough for the robot to navigate in crowded scenarios. Recently, some works utilize PointNet [15] to address robot navigation problems. PointNet first operates on each obstacle point and extracts the global obstacle features using max-pooling, which allows the change of input size. Pokle et al. [16] use PointNet to extract features from human trajectories, which helps the navigation agent know the motion of nearby people and avoid collision with them. Leiva et al. [17] utilize PointNet to extract features from laser scans, and the extracted features are concatenated with goal representation for the final decision. To ensure enough features are extracted, their PointNet adopts a three-layer DNN with 512 outputted features to process each obstacle point. Training this model requires much more computation power than fully-connected models when the number of input points is large. In their work, the trained PointNet only operates on 128 points to ensure real-time control. Moreover, when tested in real-world scenarios, the trajectories of the agent deviate far from the optimal paths (straight lines) when there are no obstacles between the goal point and the robot.

Inspired by PointNet, our model operates on each obstacle point and takes accurate positional information of these points as input. Different from PointNet, our model incorporates the goal-reaching objective into the point-wise feature extraction. Specifically, we introduce a gate mechanism to guide the network to extract obstacle features that are critical for goal-reaching. Subsequently, global features are selected from all the feature candidates. The number of global features decides the maximum number of selected points. As only a few points are used to support the final decision, we refer to these points as support points and our approach as support point-based navigation (SPN). As our model only operates on each obstacle point, the LiDAR specifications and position can be changed, provided that it can return the obstacle point position in robot frame. With high-resolution point position information as input, our model can better bypass small obstacles during navigation. To sum up, the contributions of this paper are:
\begin{itemize}
	\item A new navigation algorithm is proposed, which automatically focuses on critical obstacle points and ignores the non-critical ones.
	\item Our model can generate reliable navigation policy for robot with a new LiDAR setup without any retraining.
	\item Our model shows great potential in crowded scenarios with small obstacles when using a high-resolution LiDAR.
\end{itemize}

The remaining of this paper is organized as follows. A brief introduction of our problem is given in Section II. The proposed SPN approach is described in Section III, followed by the implementation and results in Section IV. Last, we draw the conclusions in Section V.

\section{BACKGROUND}

\subsection{Problem Formulation}
The illustration of mapless robot navigation with varied LiDAR configurations is given in Fig. 2. As shown, a circular robot is required to reach its target without colliding with any obstacles. It carries a LiDAR sensor to perceive its surroundings. The LiDAR can be described using its pose and the specifications. As shown, in the robot frame (the y-axis is robot heading direction), the pose of the LiDAR is described with $[x_l,y_l,\varphi_l]$. Besides, we define the LiDAR angular resolution as $\varphi_a$, the field of view (FOV) as $\varphi_f$ and maximum detection range as $L_{max}$. We call all the parameters above LiDAR configurations. In this paper, the robot is required to perform navigation tasks when its LiDAR configurations are changed. 

Our problem can be modeled as a sequential decision-making process. During navigation, similar to most mapless navigation problem [8], [9] and [10], the target position $s_t^g=\left[d_t^g,\varphi_t^g\right]$ in robot frame is given or directly obtained by sensors [22]. At time $t$, the input $s_t\in\mathcal{S}$ of the model contains the LiDAR readings, current robot velocities and the goal position. The dimension of $s_t$ may change with different LiDAR configurations. The action $a_t\in\mathcal{A}$ of the robot comprises the commands of linear and angular velocities. Given $s_t$, the robot takes action $a_t$ controlled by the current policy $\pi$. With new observations received at time $t+1$,  the robot updates its input $s_{t+1}$ and obtains a reward $r_t(s_t,a_t,s_{t+1})$ from the reward function. The objective of this paper is to find an optimal policy $\pi^\ast$ that maximizes the discount total rewards $G_{t}=  \Sigma  _{ \tau=t}^{T} \gamma ^{ \tau-t}r_{ \tau}$, where $\gamma\in[0,1)$ is a discount factor.
\begin{figure}[tb]
	\centering
	\includegraphics[width=0.9\linewidth]{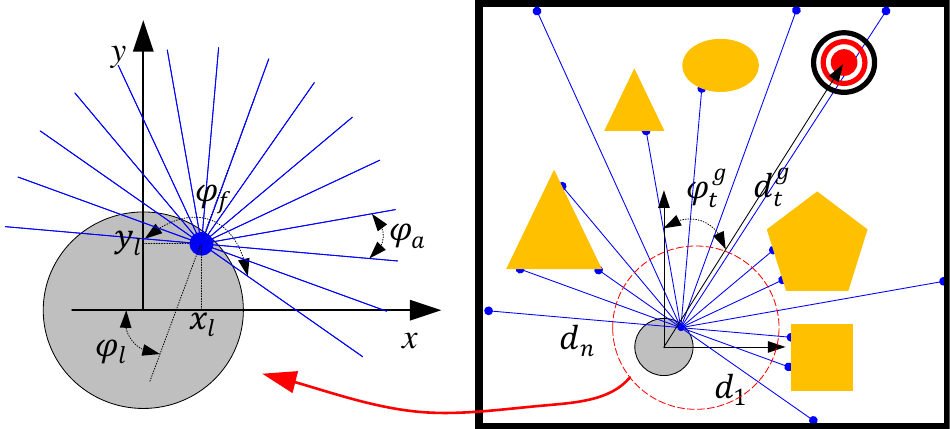}
	\caption{Illustration of robot navigation problem with varied LiDAR configurations.}
	\label{fig:2}
\end{figure}

\section{APPROACH}

We use DRL to learn a DNN model $\pi_{\theta}$ to approximate the optimal policy $\pi^\ast$. The DRL algorithm used in this paper is soft actor critic (SAC) [18]. It contains an actor network for approximating the policy and three critic networks (one value network, and two $Q$ networks) for approximating expected total returns. The structures for these networks are given in Fig. 3. As shown, the actor network takes point-wise information as input, while the critic networks share a similar structure with our previous work [19], which take 1D laser scans as input.
\subsection{Actor Network} 
\subsubsection{Input representation}
To learn DNN-based navigation controller from laser scans, previous works usually format the laser scans as a one-dimensional vector $\mathbf{d}=\left[d_1,\cdots,d_n\right]$, where $d_i$ denotes the $i$-th scanned distance value and $n$ ($n=1080$ in this paper) is the number of laser beams. The direction of each laser beam is included by its location in $\mathbf{d}$. However, such representations restrict the trained model from handling the data obtained by LiDARs with different FOVs or angular resolutions. To address this problem, in this paper,  the laser scans are represented by an $n$-element set $\mathcal{P}=\left\{\mathbf{p}_i| i=1,\cdots,n\right\}$, where $\mathbf{p}_i=[{\sin{\alpha_i}/d}_i, {\cos{\alpha_i}/d}_i]$ and $\alpha_i$ is the relative angle of the obstacle point in robot frame. Same as our previous work [19], we use the reciprocal of distance to highlight the small-distance values in laser scans. According to such representation, the position of each obstacle point is explicitly described, and hence there is no ordering requirement for elements in $\mathcal{P}$.
\begin{figure}
	\begin{minipage}{.45\linewidth}
		\begin{subfigure}{\linewidth}
			\centering
			\includegraphics[width=0.95\linewidth]{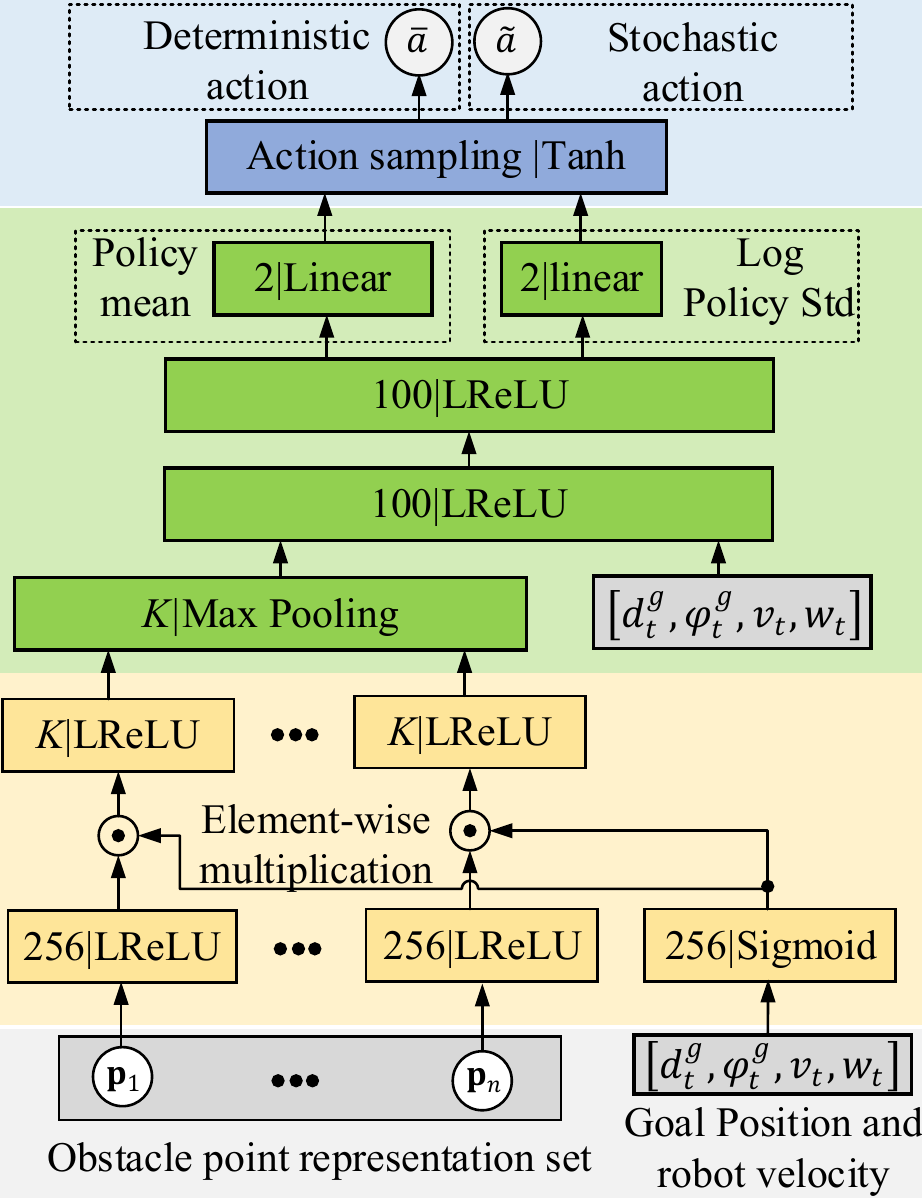}
			\caption{Actor network}
			\label{fig:sub3}
		\end{subfigure}
	\end{minipage}%
	\begin{minipage}{.50\linewidth}
		\begin{subfigure}{\linewidth}
			\centering
			\includegraphics[width=0.9\linewidth]{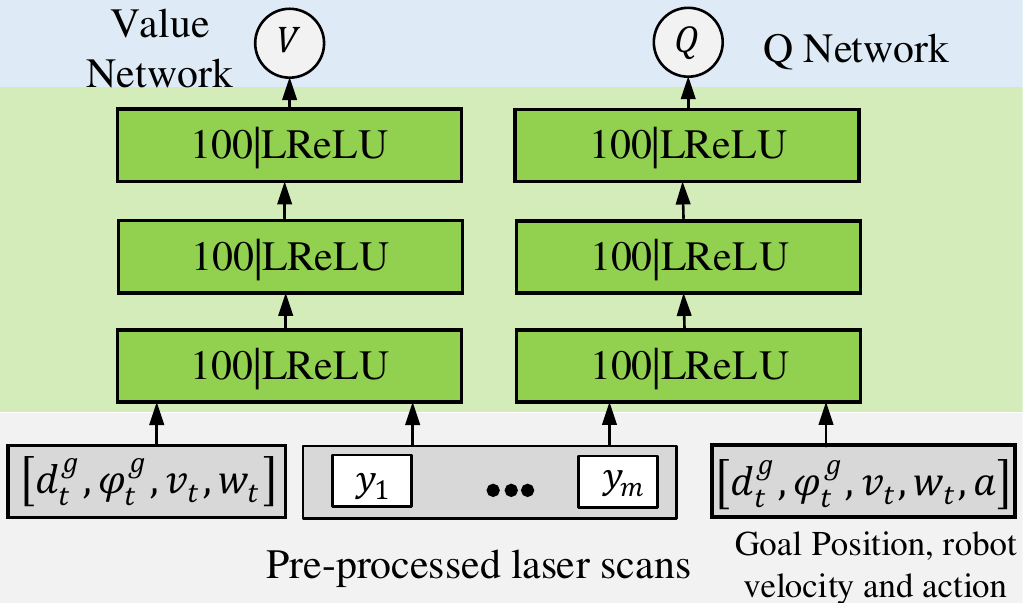}
			\caption{Critic networks for SPN}
			\label{fig:sub1}
		\end{subfigure}\\[1ex]
		\begin{subfigure}{\linewidth}
			\centering
			\includegraphics[width=0.9\linewidth]{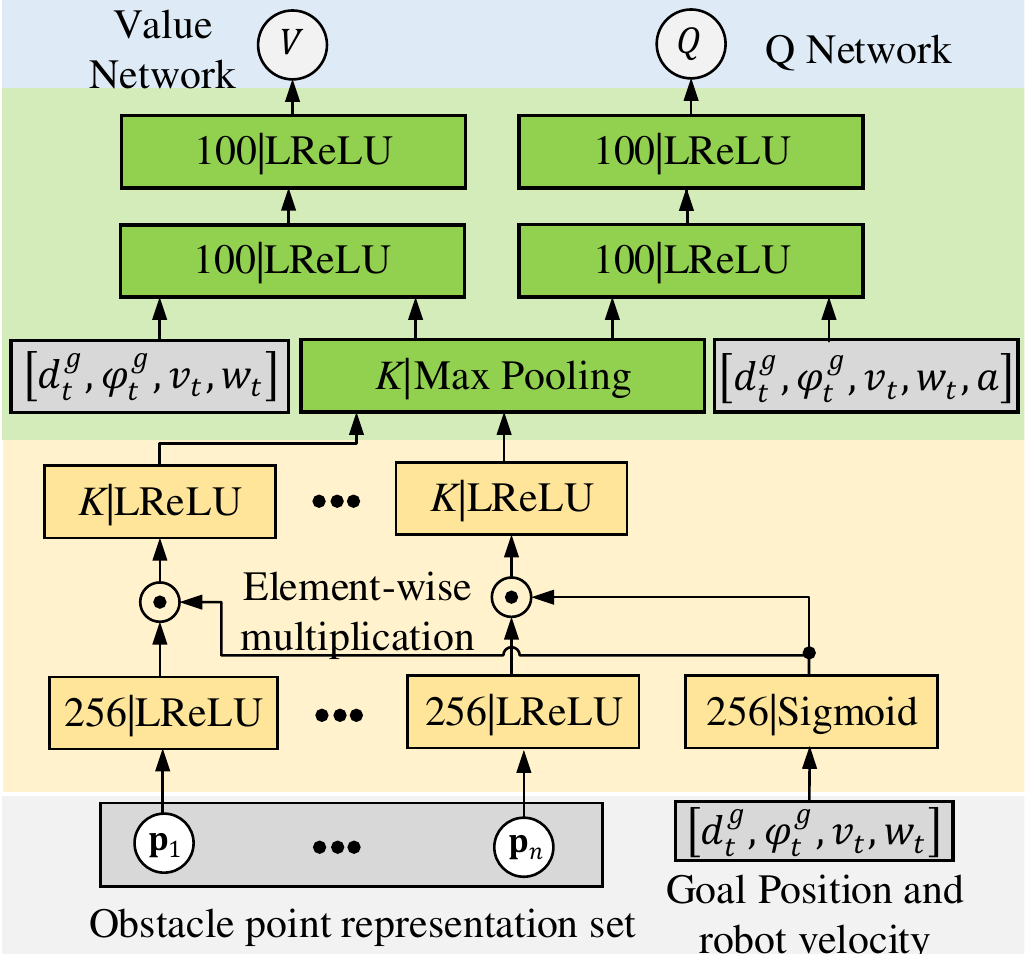}
			\caption{Critic networks for SPN-v2}
			\label{fig:sub2}
		\end{subfigure}
	\end{minipage}

	\caption{Network structures of the actor network and critic networks.}
	\label{fig:test}
\end{figure}
\subsubsection{Goal-directed point-wise feature extraction}The shadow layers (labeled with orange in Fig. 3(a)) of actor network operate on each point representation $\mathbf{p}_i$ individually. As shown, each obstacle point is fed into a dense layer and activated by LReLU (Leaky Rectified Linear Unit). Meanwhile, the relative position of the goal $s_t^g=\left[d_t^g,\varphi_t^g\right]$ and current robot velocity $\left[v_t,\omega_t\right]$ are fed into another dense layer and activated by a sigmoid function. Subsequently, we apply element-wise multiplication to the two features as follows,
\begin{equation}
\begin{aligned}
\mathbf{h}\left(\mathbf{p}_{i}\right)=\mathrm{lrelu}\left(\mathbf{W}_{1}\mathbf{p}_{i}^{\mathsf{T}}+\mathbf{b}_{1}\right)\odot\mathrm{sigm}\left(\mathbf{W}_{2}\mathbf{g}^\mathsf{T}+\mathbf{b}_{2}\right)
\end{aligned}
\end{equation}
where $\mathbf{W}_{1}\in\mathbb{R}^{H\times 2}$,$\mathbf{W}_{2}\in\mathbb{R}^{H\times 4}$,$\mathbf{b}_\mathbf{1},\mathbf{b}_\mathbf{2}\in\mathbb{R}^{H\times 1}$ are network weights and bias of the aforementioned two layers, $\odot$ dentotes the element-wise multiplication, $\mathrm{lrelu}\left(\cdot\right)$ denotes the LReLU function, $\mathrm{sigm}\left(\cdot\right)$ denotes the sigmoid function and $\mathbf{g}=\left[d_t^g,\varphi_t^g,v_t,\omega_t\right]$. The sigmoid function $\mathrm{sigm}\left(\cdot\right)\in\left(0,1\right)$) here serves as a gate function that helps reserve important obstacle features and remove non-critical ones given the goal position and current velocity. 
\subsubsection{Global feature selection and policy calculation}
The following network structure is similar to PointNet [14]. The filtered feature representation $\mathbf{h}\left(\mathbf{p}_{i}\right)$ is fed into another dense layer with $K$ output neurons, followed by a max-pooling operation. Specifically, each obstacle point is finally represented as a $K$-dimensional feature vector $\mathbf{f}\left(\mathbf{p}_{i}\right)\in\mathbb{R}^{K}$. For each feature channel $j\left(1\leq j\leq K\right)$, the maximum feature value is chosen as the global feature $\mathbf{g}_j$ for supporting final decision, which is as follows,
\begin{equation}
\begin{aligned}
\mathbf{g}_j = \mathbf{f}_j\left(\mathbf{p}^{*}\right)=\max_{1\leq i\leq n}\mathbf{f}_j\left(\mathbf{p}_{i}\right)
\end{aligned}
\end{equation}
As point $\mathbf{p}^{*}$ contributes to this feature, we call $\mathbf{p}^{*}$ as support point in this paper. Through max-pooling, the global features of all obstacle points are extracted. The number $K$ of global features is a key factor, which decides the maximum number of obstacle points used for the final decision.The selected obstacle features, combined with goal position and current velocity, are fed into a two-layer DNN for policy calculation. As squashed Gaussian policy is used in this paper, we refer to the sampled stochastic action (used in training) as $\widetilde{a}$ and the deterministic action (used in testing) as $\bar{a}$.

\subsection{Critic Networks}
As critic networks are not used in testing, their input representation and the network structure are not necessarily similar to the actor network. In this paper, we adopted two kinds of structures for our critic networks. The first structure of critic network is similar to [8] and [19]. As shown in Fig. 3(b), all critic networks are fully-connected three-layer DNNs. To reduce the input dimension and the number of network parameters, the original laser scans are compressed into $m$ ($m=36$ in this paper) values by 1D minimum down-sampling. Similar to actor network, we choose the reciprocal of the down-sampled value as the input representation $y_i$, which is computed as,
\begin{equation}
\begin{aligned}
y_i=\frac{1}{\min{\left(d_{i\cdot k+1},d_{i\cdot k+2},\ \cdots,d_{i\cdot k+k}\right)}}
\end{aligned}
\end{equation}
where $i$ is the index of the down-sampled laser scans and $k$ ($k=30$ in this paper) is the length of down-sampling window. The down-sampled representations of laser scans, together with relative goal position and robot velocity, are fed into the value network to approximate the value function. The double $Q$ networks (only one Q network is plotted in Fig. 3(b)) share the same structure as the value network except that it also takes the agent's action as input. The second kind of critic networks adopts the same input and network structure as the actor network. To be computation-efficient, as shown in Fig. 3(c), the Value network and double $Q$ networks share the same goal-directed point feature extraction layers. In order to distinguish easily, we name the former model as SPN and the latter one as SPN-v2.
\subsection{Training Approach}
The agent's learning objective is to reach a given goal point in the shortest time-steps without colliding with the obstacles. To guide the learning process, the reward function contains a positive part $r_s$ for encouraging goal-reaching and a negative part $r_c$ for punishing collision. As the above rewards are too sparse to be received during training, similar to [9], an additional dense part is adopted as follows,
\begin{equation}
\begin{aligned}
r_t = 
\begin{cases}
r_{s}, & \text{if success}\\
r_{c}, & \text{if crash}\\
c_{1} \left( d_{t}^{g}-d_{t+1}^{g} \right) + c_{2}, & \text{otherwise}.
\end{cases}
\end{aligned}
\end{equation}
where $c_1$ is a scaling constant, and $c_2$ is a small negative constant for punishing staying still (some works [10] may not have this time-penalty term). 

In SAC [18], the expected total return is regularized by the policy entropy and as follows,:
\begin{equation}
J\left(\pi\right)=\mathbb{E}_{\pi}\left[G_{t=0}+\Sigma_{t=0}^{T}\gamma^{t}\alpha \mathcal{H} \left(\pi\left(\vert s_{t}\right)\right)\right].
\end{equation}
where $\mathcal{H}\left(\pi({\cdot|s}_t)\right)=-\int_{\left|\mathcal{A}\right|}{\pi\left({a|s}_t\right)\log{\pi\left({a|s}_t\right)\mathrm{d}a}}$ is policy entropy, and $\alpha$ is a weighing factor. During training, the critic networks are optimized to approximate $J\left(\pi\right)$ using the Bellman equation, and the actor network learns a policy that maximizes the $Q$ values approximated by the $Q$ network. 

\section{IMPLEMENTATION AND TESTS}
\subsection{Model Training}
We train our model in four simulated scenarios using ROS Stage [20] in a workstation with NVIDIA Quadro P5000 GPU. As shown in Fig. 4, the simulated differential robot is a circular robot with a radius of $0.2\mathrm{m}$. It maximum linear velocity is $0.5\mathrm{m/s}$ and maximum angular velocity is $\pi/2\mathrm{rad/s}$ A LiDAR is mounted on the robot center. Its FOV is ${360^{\circ}}$, angular resolution is $0.33^{\circ}$ (1080 laser beams) and scanning range is $5\mathrm{m}$. The four training scenarios are ordered (left to right) by their training difficulty: Env(0) is the simplest, and Env(3) is the hardest. To accelerate training, as shown in Fig. 5, we adopt curriculum learning as follows. At beginning, the selection probability of the four scenarios {Env(0), Env(1), Env(2), Env(3)} is $\left[0.7,0.1,0.1,0.1\right]$. When the agent's success rate in Env($i$) reaches 0.9 in the past 50 episodes, the selection probability of the next scenario Env($i+1$) exchanges with the selection probability of Env($i$). When the selection probability of Env(3) reaches 0.7, we keep the selection probability of each scenario until the end of training. 
\begin{figure}[t]
	\centering
	\includegraphics[width=0.98\linewidth]{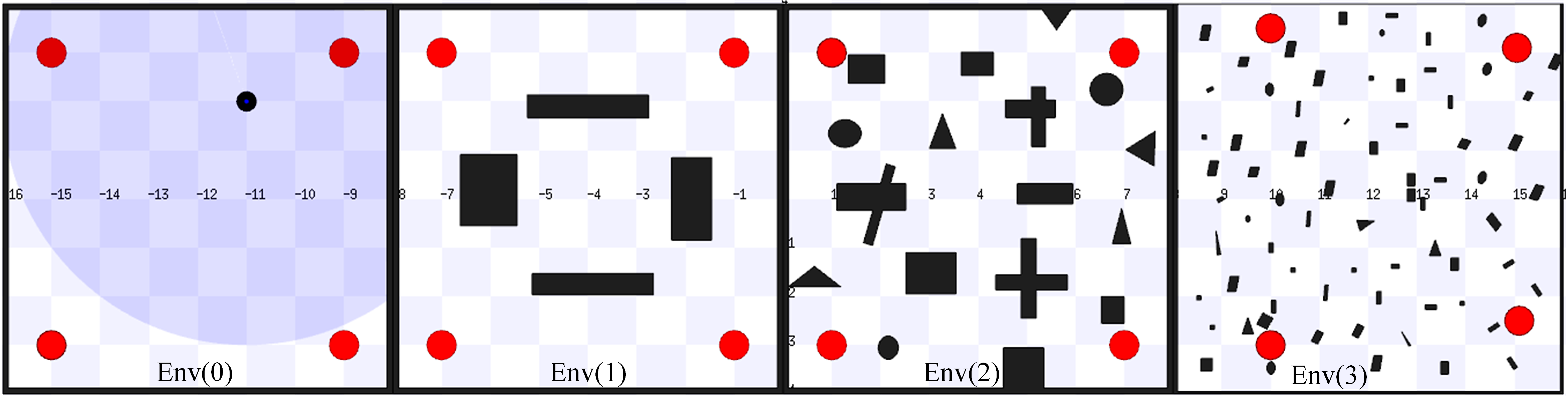}
	\caption{The simulated scenarios used for training.}
	\label{fig:4}
\end{figure}
\begin{figure}[t]
	\centering
	\includegraphics[width=0.98\linewidth]{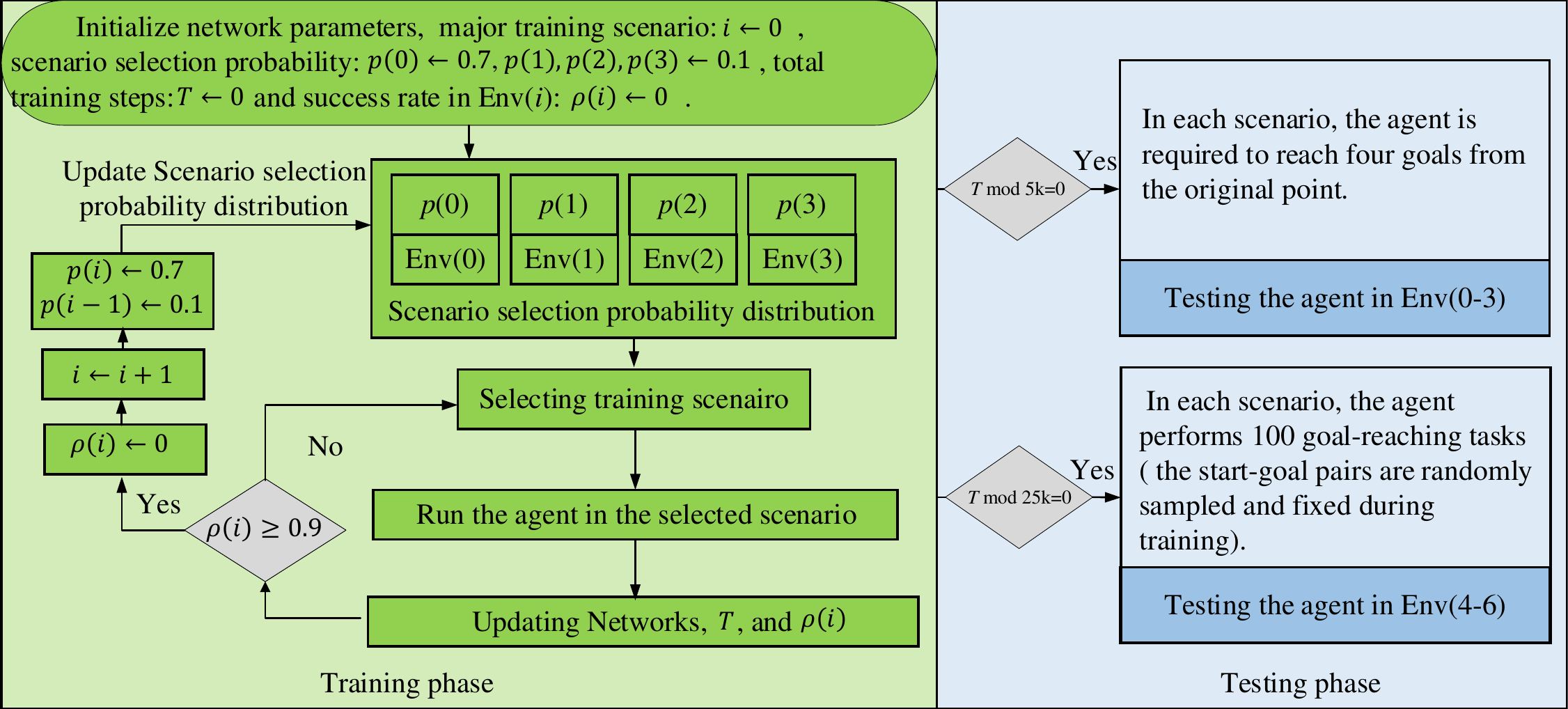}
	\caption{Flowchart of training and testing phases.}
	\label{fig:4}
\end{figure}
\begin{figure}[t]
	\centering
	\includegraphics[width=0.8\linewidth]{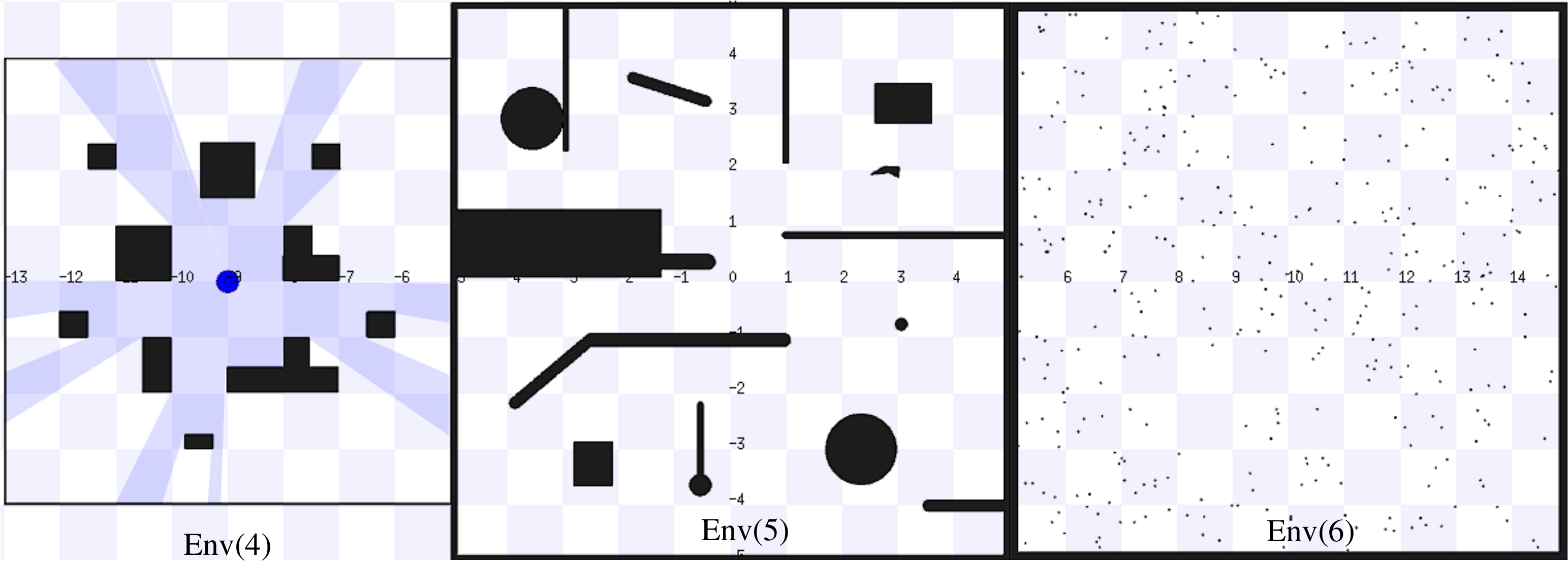}
	\caption{The simulated scenarios used for testing.}
	\label{fig:4}
\end{figure}
\begin{figure}[htbp]
	\centering
	\includegraphics[width=0.99\linewidth]{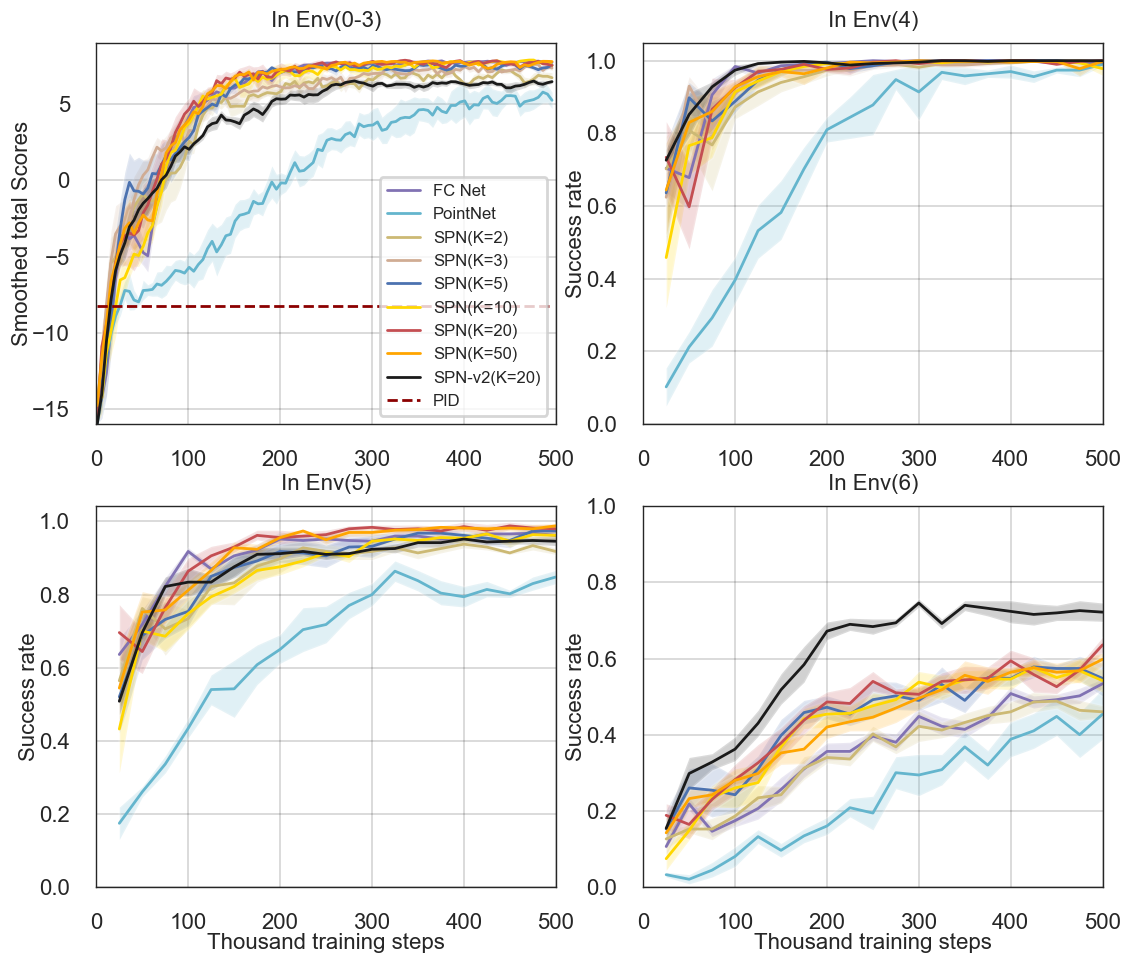}
	\caption{Averaged learning curves of different models. The solid lines represent the averaged scores or success rates, and the translucent areas indicate the variance of scores or success rates in five runs.}
	\label{fig:5}
\end{figure}
\begin{figure*}[htbp]
	\centering
	\begin{subfigure}{0.24\linewidth}
		\centering
		\centerline{\includegraphics[width=0.98\linewidth]{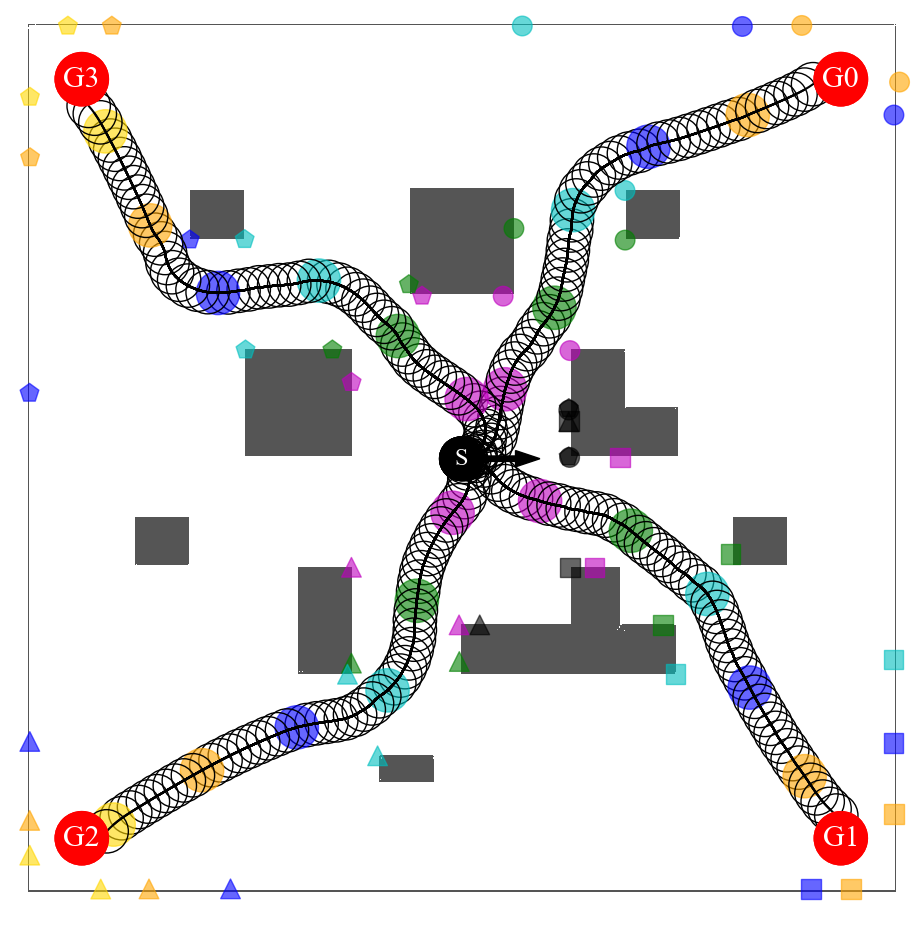}}  
		\caption{$K=2$}
		\label{fig:6a}
	\end{subfigure}
	\hfill
	\begin{subfigure}{0.24\linewidth}
		\centering
		\centerline{\includegraphics[width=0.98\linewidth]{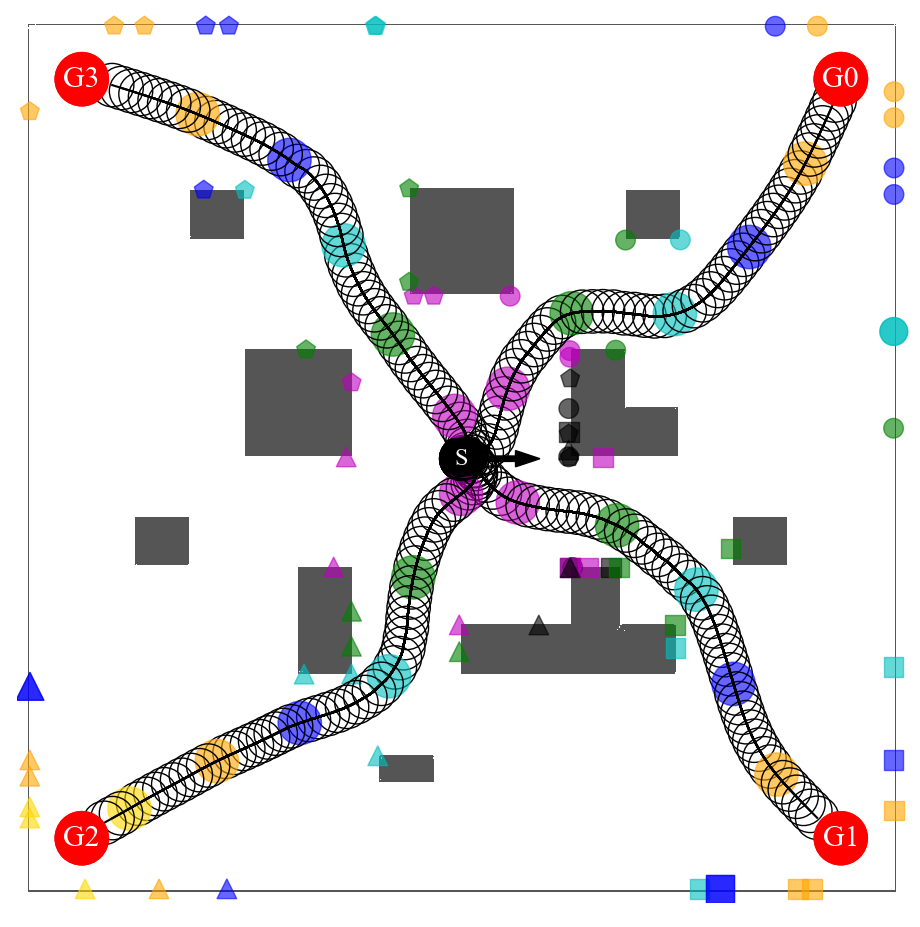}}
		\caption{$K=3$}
		\label{fig:6b}
	\end{subfigure}
	\hfill
	\begin{subfigure}{0.24\linewidth}
		\centering
		\centerline{\includegraphics[width=0.98\linewidth]{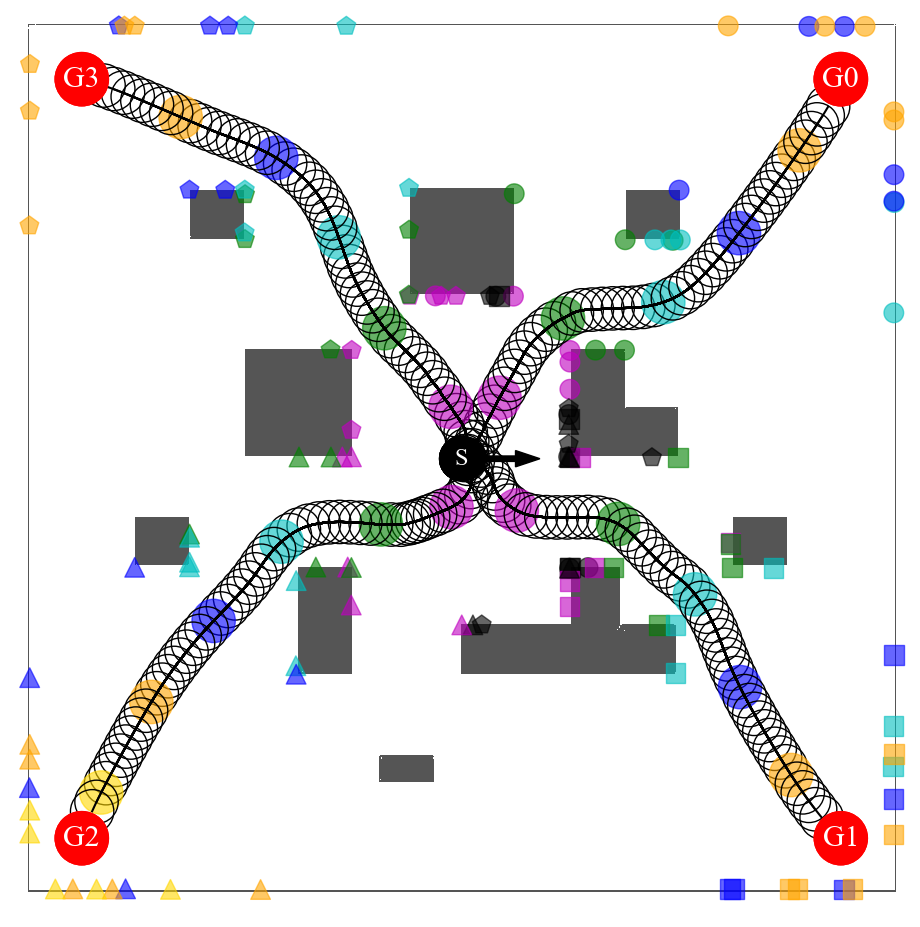}}
		\caption{$K=5$}
		\label{fig:6b}
	\end{subfigure}
	\hfill
	\begin{subfigure}{0.24\linewidth}
		\centering
		\centerline{\includegraphics[width=0.98\linewidth]{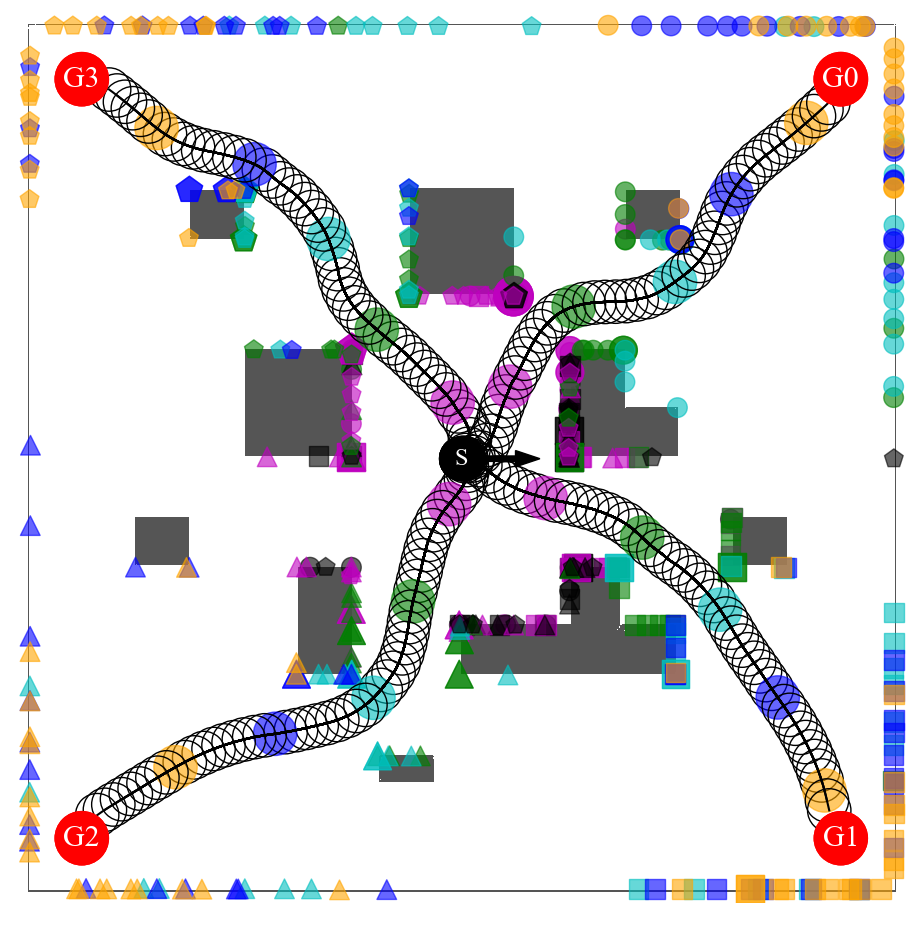}}
		\caption{$K=20$}
		\label{fig:6b}
	\end{subfigure}
	\caption{Support point visualization for SPN models with (a) $K=2$, (b) $K=3$, (c) $K=5$ and (d) $K=20$ in simulated scenairo Env(4) [9]. The support-point markers used for the four tasks are circle (S-G0), square (S-G1), triangle (S-G2), and pentagon (S-G3).}
	\label{fig:6}
\end{figure*}
In each training episode, the robot is spawned in a scenario arranged by the selection probability. It starts from a randomly chosen obstacle-free point and aims to reach a goal point (not rendered). In the first 100 episodes, the agent is controlled by a PID controller [9]. Afterwards, it takes the stochastic action sampled from squashed Gaussian policy.The episode will end when the robot reaches the target, crashes or time out(the maximum time steps per episode is 400). The control command sent to the robot is 0.1s. We train the model 500k steps and test the model every 5k training steps in training scenarios. When tested in training scenarios, in each scenario, the agent is required to reach four goals (labeled with red circles) from the original point with the deterministic policy. To validate the genealization capability of our agent, every 25k steps, we test the agent in three unseen scenarios: Env(4)[9], Env(5)[10] and Env(6). In each scenario, one hundred initial and goal points are randomly sampled and fixed during the whole training process. During testing, the agent needs to perform 100 goal-reaching tasks in each scenario.

To investigate the effect of the number $K$ of global features, we train six SPN models with $K\in\left\{2, 3, 5, 10, 20, 50\right\}$. Besides, we also train SPN-v2($K$=20) to investigate the effect of using different architectures of the critic networks. We compare our SPN models with 1) FC Net, which takes 1D LiDAR data as input, and its input and structure are the same as the value network used in our SPN model; 2) PointNet, which takes raw point coordinates as input, and its network structure is similar with SPN-v2($K$=20), except the goal-directed feature extraction process is changed back to point feature extraction. Each model is trained five times using different random seeds. As the dense part in the reward function cannot represent the real navigation objective, the metric used to evaluate the model's performance in training scenarios is similar to [9] and is represented by a score $S$ function as follows,
\begin{equation}
\begin{aligned}
S = 
\begin{cases}
1-2T_s/T_{max}, & \text{if success}\\
-1, & \text{otherwise}.
\end{cases}
\end{aligned}
\end{equation}
where $T_s$ is the total number of steps spent on navigation, and $T_{max}$ ($T_{max}=400$ in this paper) is the maximum time steps per testing episode. As shown, the robot will get the same negative score when its ending status is collision or time out. The metric used to evaluate the model's performance in testing scenarios is the success rate of 100 runs in each scenario. The learning curves are given in Fig. 7. As shown, our SPN models ($5\leq K\leq50$) converge to high scores in the training scenarios and slightly outperform the FC Net. It is noteworthy that our SPN model performs better than FC Net in Env(5) and achieves much higher success rate in Env(6). It indicates the SPN model can generalize better navigation policies in scenarios with small and dense obstacles than the FC Net. Notably, although our SPN-V2 model achieves a relatively low score in the training scenario, it demonstrates the best performance in Env(6). The PointNet learns much slower than the other models, which shows that the input representation and goal-directed feature extraction used in SPN-v2 can greatly improve its learning performance. As SPN($K$=20) performs the best in Env(6), in the remaining of this paper, we use SPN($K$=20) for testing.

\subsection{Support Points Visualization}
To better understand the effect of the support points, we visualize them in Env(4) shown in Fig. 8. This map is an $8\times8m^2$ square room, which is considered a complex scenario in [9]. SPN($K$=2), SPN($K$=3), SPN($K$=5), and SPN($K$=20) are used for testing. As shown in Fig. 8, the robot is initialized at the center point (black) and facing right (black arrow). It is required to reach four goal-points (G0-G3) from the initial state respectively. We plot the support points selected by our model every 20 control steps for each task. As shown, the support points are marked with different shapes according to their goal, and their colors are consistent with the colors of the footprints of the robot. As several features may come from the same obstacle point, we amplify the size of the support point marker by $j$ times if $j$ features are extracted from this point. Notably, at beginning of each task, the LiDAR readings are the same, but the support points are different (see black markers). This shows the learned representations and corresponding policy from the LiDAR readings are goal-conditioned, i.e., the model only reserves obstacle features that are valuable for goal-reaching. Moreover, our model can accurately capture the corner points that may affect its navigation in the near future and ignore the nearby points that have been passed or in the opposite direction of its goal.

\subsection{Performance Evaluation in Simulated Scenarios}
To evaluate the performance of our model, robots with seven LiDAR configurations containing different angular resolutions, FOVs, and LiDAR positions are tested in three simulated scenarios given in Fig. 6. For simplicity, we label each LiDAR setup with \{$\varphi_f\left(^\circ\right)|\varphi_a \left(^\circ\right)| L_{max}\left(\mathrm{m}\right)|y_l\left(\mathrm{m}\right)$\}, where the first three parameters represent the specification of the LiDAR and the last term represents the LiDAR position (see Fig. 2 for parameter definitions). The other two position parameter $x_l$ and $\varphi_l$ are both zero. Three of the LiDAR setups (${``180|20|10|0"}$, ${``240|0.47|5.6|0"}$, ${``270|0.25|30|0"}$) are used in real-world experiments in [8], [9] and [10], respectively. The training setup ${``360|0.33|5|0"}$ and a degraded configuration ${``360|10|5|0"}$ are also tested to investigate the influence of changed angular resolution. We also change the installtion position of the Lidar used in [8] into ${``180|20|10|0.15"}$ and ${``180|20|10|-0.15"}$ to investigate influence of Lidar position.

Three approaches, i.e., SPN($K$=20), SPN-v2($K$=20), and FC Net, are tested and compared. Some LiDAR configurations may not meet the input format requirement of the FC Net. To address this problem, we treat the unscanned areas as free space and pad the missing data with maximum scanning distance of the lidar. In each scenario, the robot performs 100 navigation tasks with randomly generated initial and goal positions. Some trajectories of the agent are plotted in Fig. 9, and the average ratio of ending status (success, crash, or time out) is shown in Fig. 10. As shown, all robots with high-resolution LiDARs ($\varphi_a \leq1{^\circ}$) can achieve around 90\% success rate in three scenarios with SPN and SPN-v2, which demonstrates a high generalization performance of our model. Notably, our SPN and SPN-v2 models outperform FC Net under all tested LiDAR configurations. When the Lidar angular resolution becomes lower, the difference of the success rate between SPN and FC Net increases, which indicates our SPN models are more suitable to address tasks with changed LiDAR configurations. 
\begin{figure*}[htbp]
	\centering
	\begin{subfigure}{.24\linewidth}
		\centering
		\centerline{\includegraphics[width=.95\linewidth]{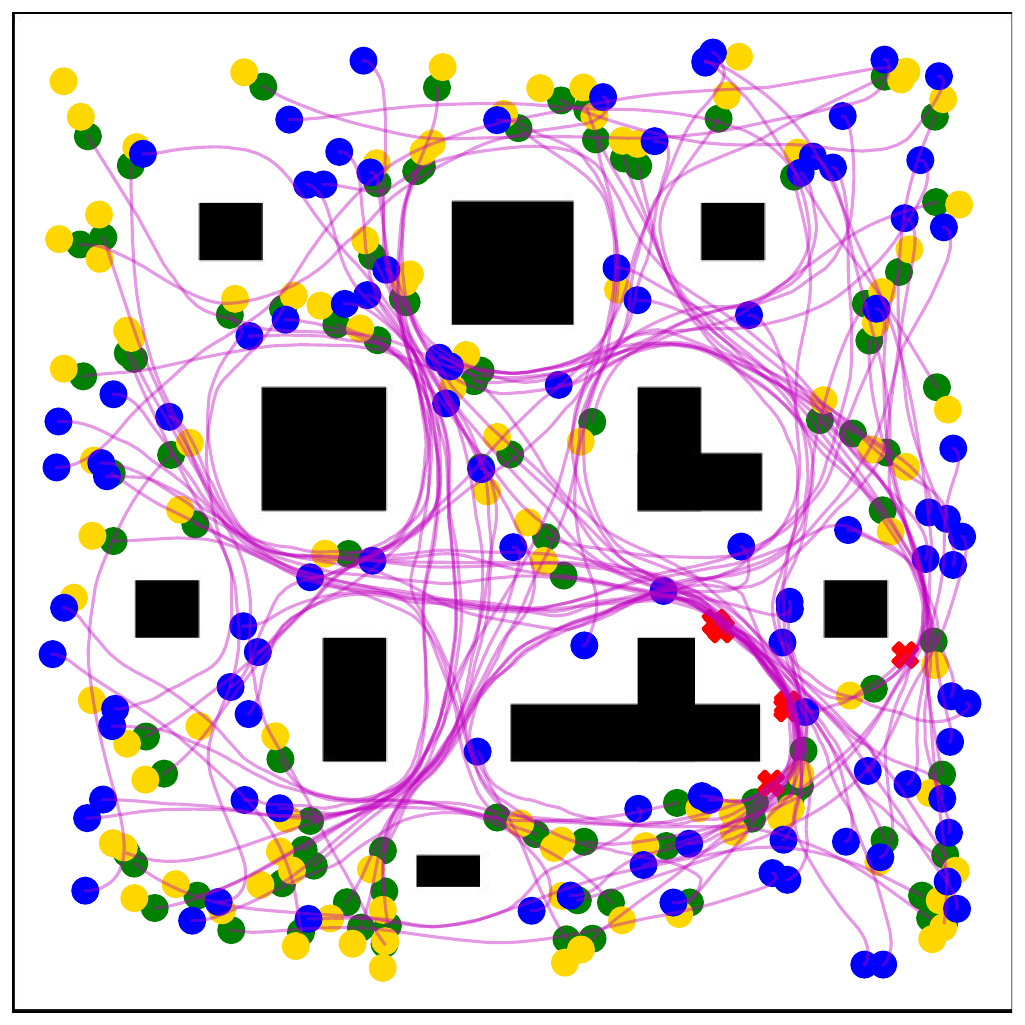}}  
		\caption{SPN: ${180|20|10|0}$}
		\label{fig:7a}
	\end{subfigure}
	\hfill
	\begin{subfigure}{.24\linewidth}
		\centering
		\centerline{\includegraphics[width=.95\linewidth]{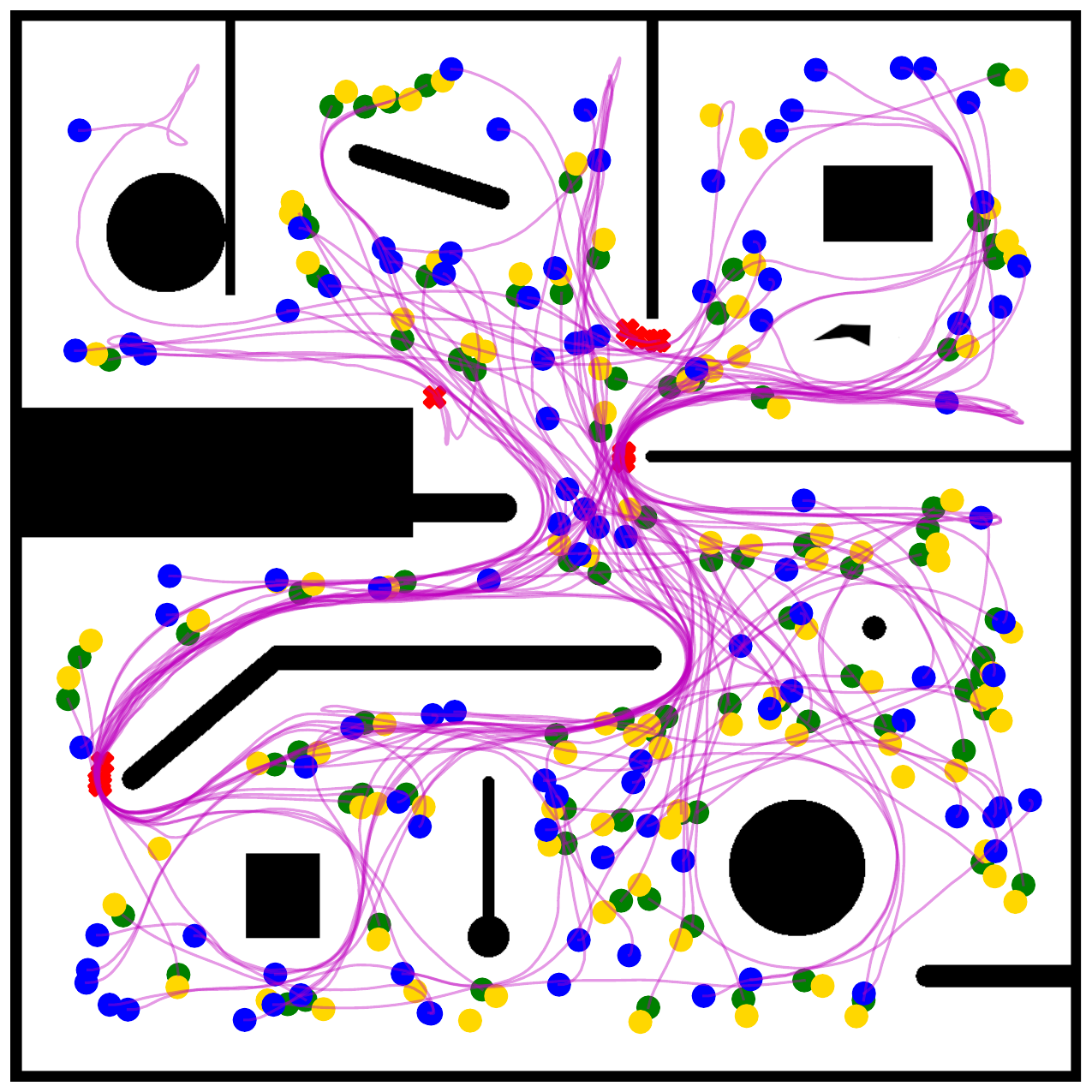}}
		\caption{SPN: ${180|20|10|0.15}$}
		\label{fig:7b}
	\end{subfigure}
	\hfill
	\begin{subfigure}{.24\linewidth}
		\centering
		\includegraphics[width=.95\linewidth]{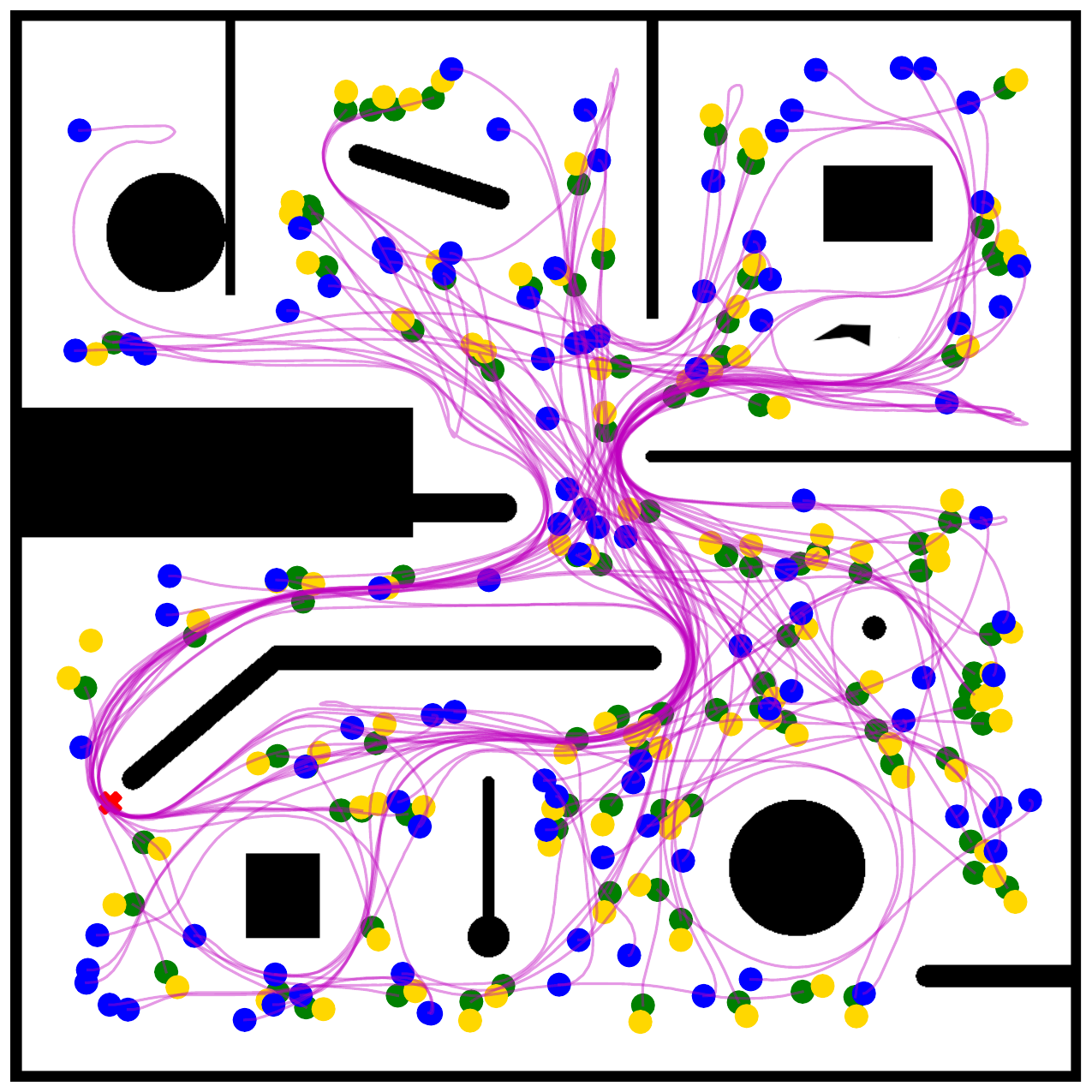}  
		\caption{SPN: ${360|10|5|0}$}
		\label{fig:7c}
	\end{subfigure}
	\hfill
	\begin{subfigure}{.24\linewidth}
		\centering
		\includegraphics[width=.95\linewidth]{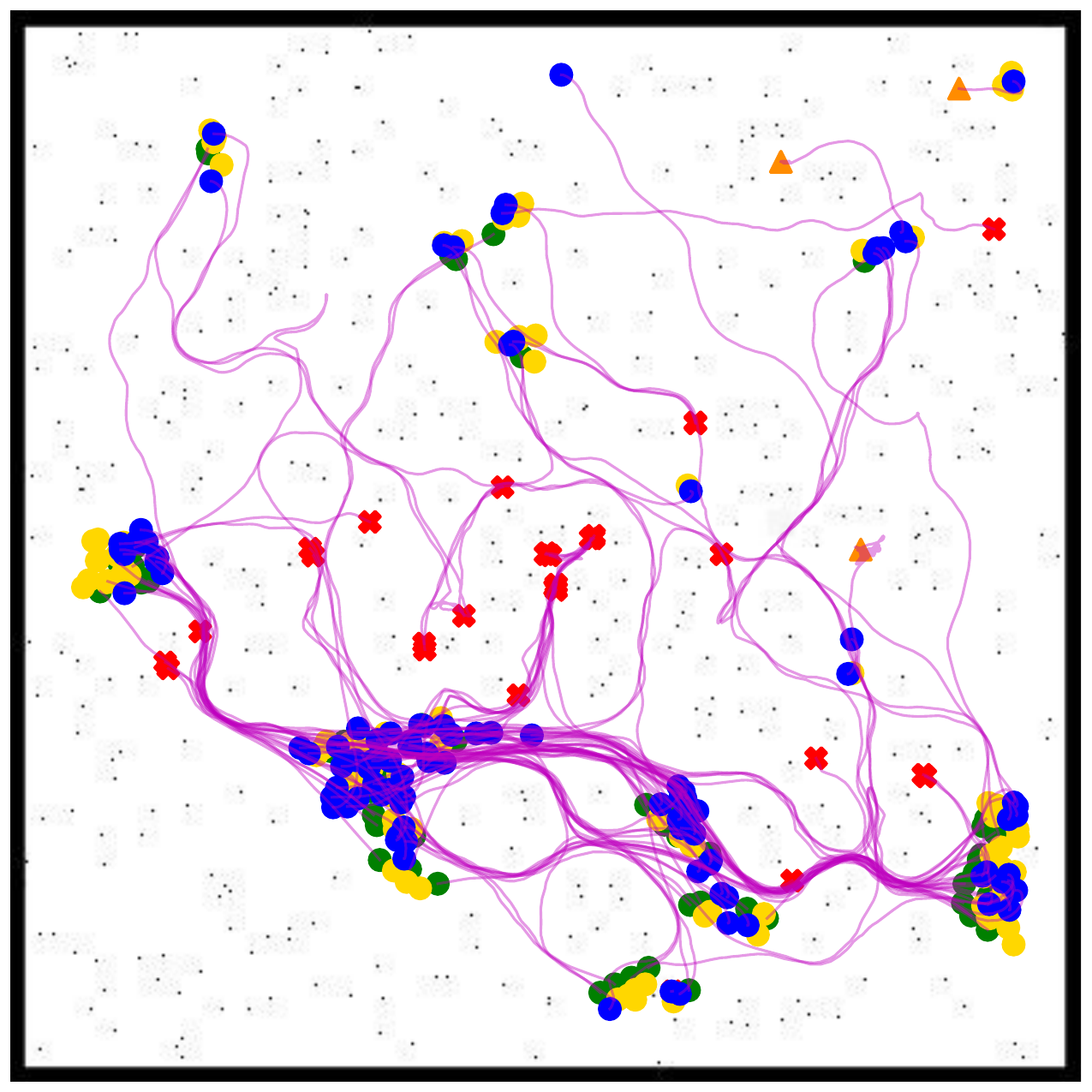}  
		\caption{SPN: ${270|0.25|30|0}$}
		\label{fig:7d}
	\end{subfigure}
	\newline
	\begin{subfigure}{.24\linewidth}
		\centering
		\centerline{\includegraphics[width=.95\linewidth]{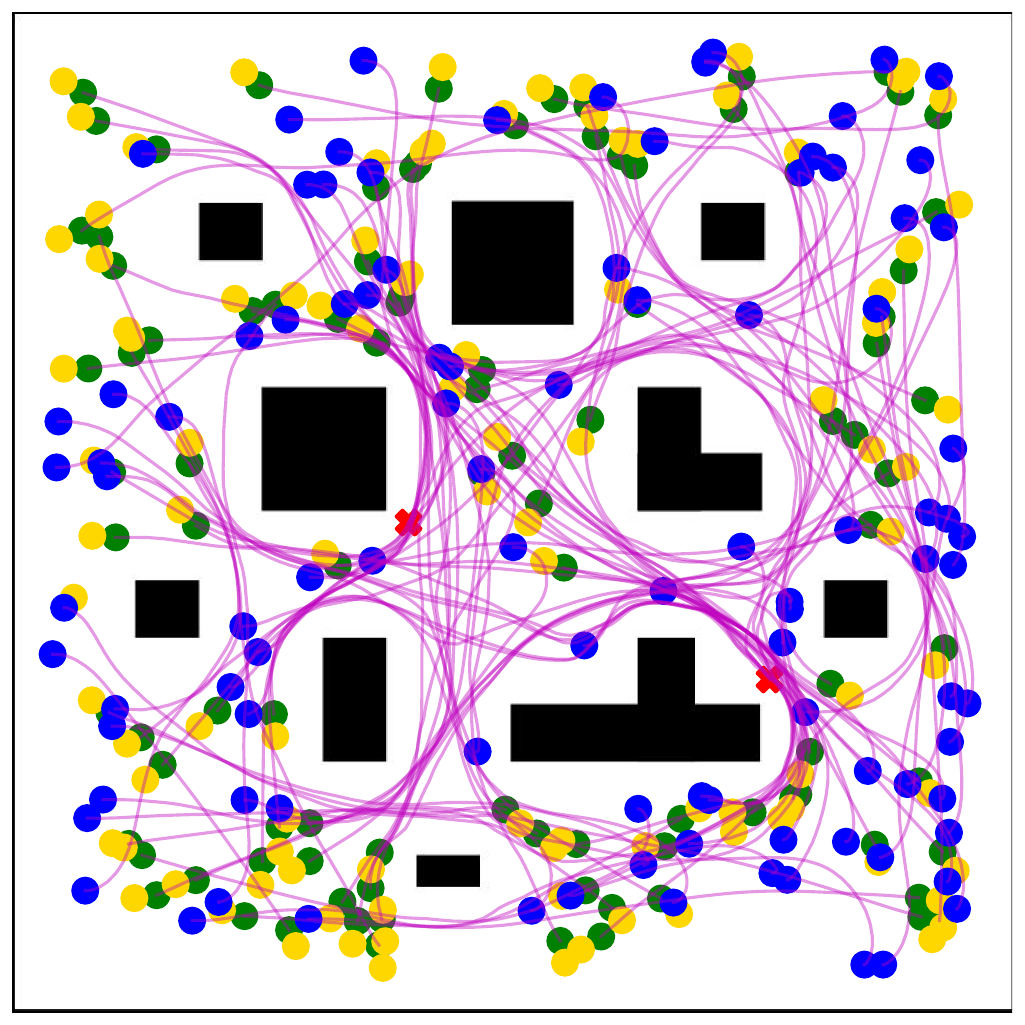}}  
		\caption{SPN-v2: ${180|20|10|0}$}
		\label{fig:7a}
	\end{subfigure}
	\hfill
	\begin{subfigure}{.24\linewidth}
		\centering
		\centerline{\includegraphics[width=.95\linewidth]{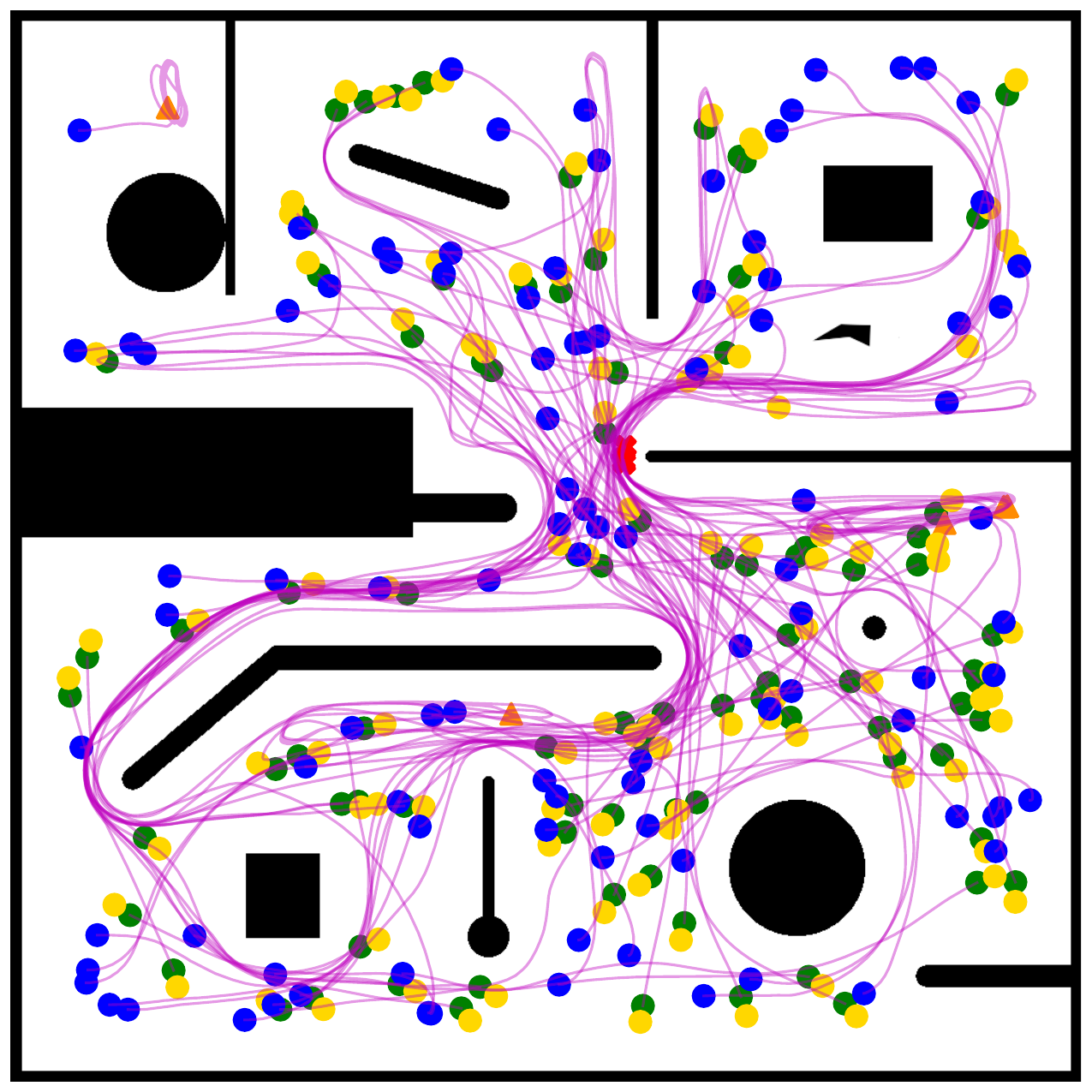}}
		\caption{SPN-v2: ${180|20|10|0.15}$}
		\label{fig:7b}
	\end{subfigure}
	\hfill
	\begin{subfigure}{.24\linewidth}
		\centering
		\includegraphics[width=.95\linewidth]{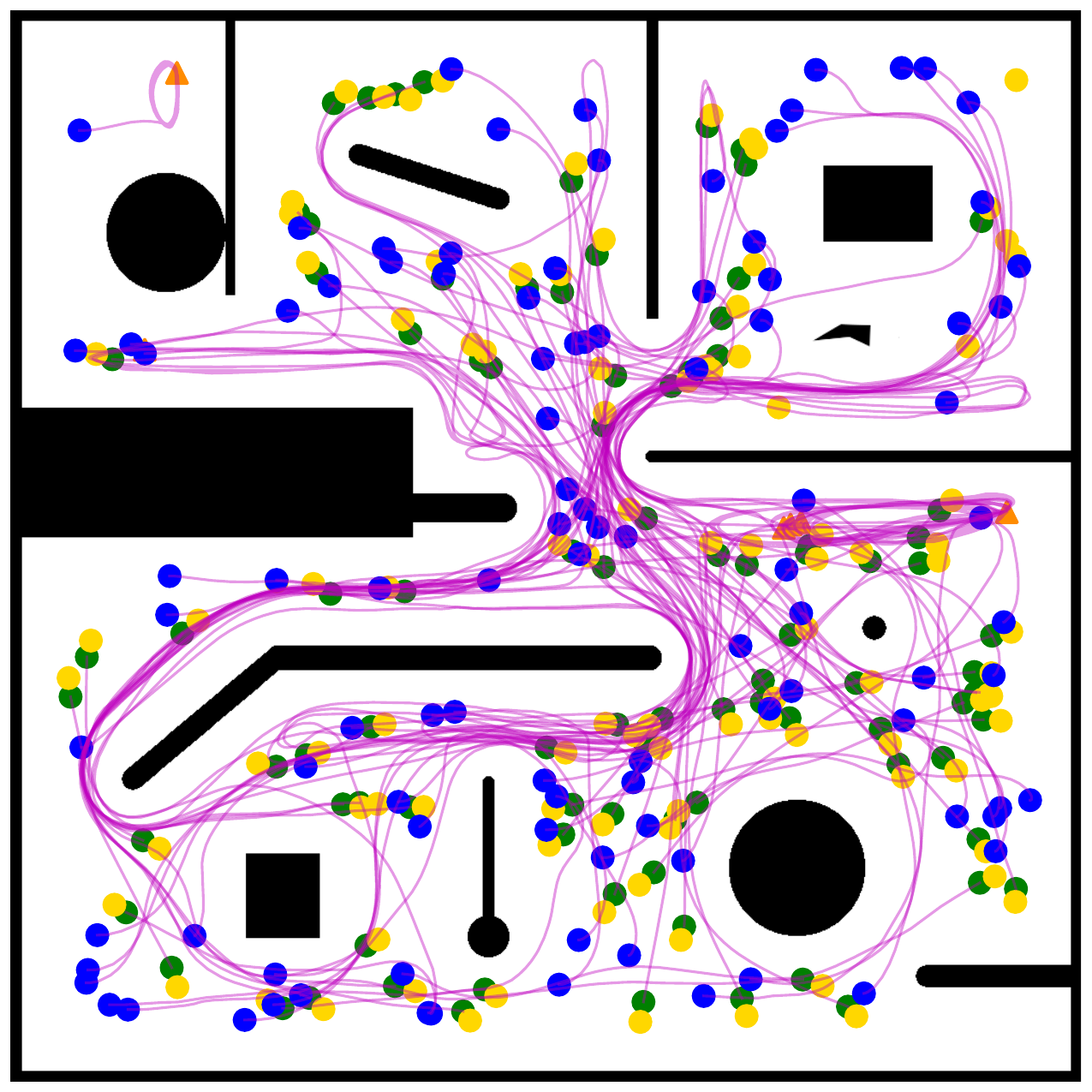}  
		\caption{SPN-v2: ${360|10|5|0}$}
		\label{fig:7c}
	\end{subfigure}
	\hfill
	\begin{subfigure}{.24\linewidth}
		\centering
		\includegraphics[width=.95\linewidth]{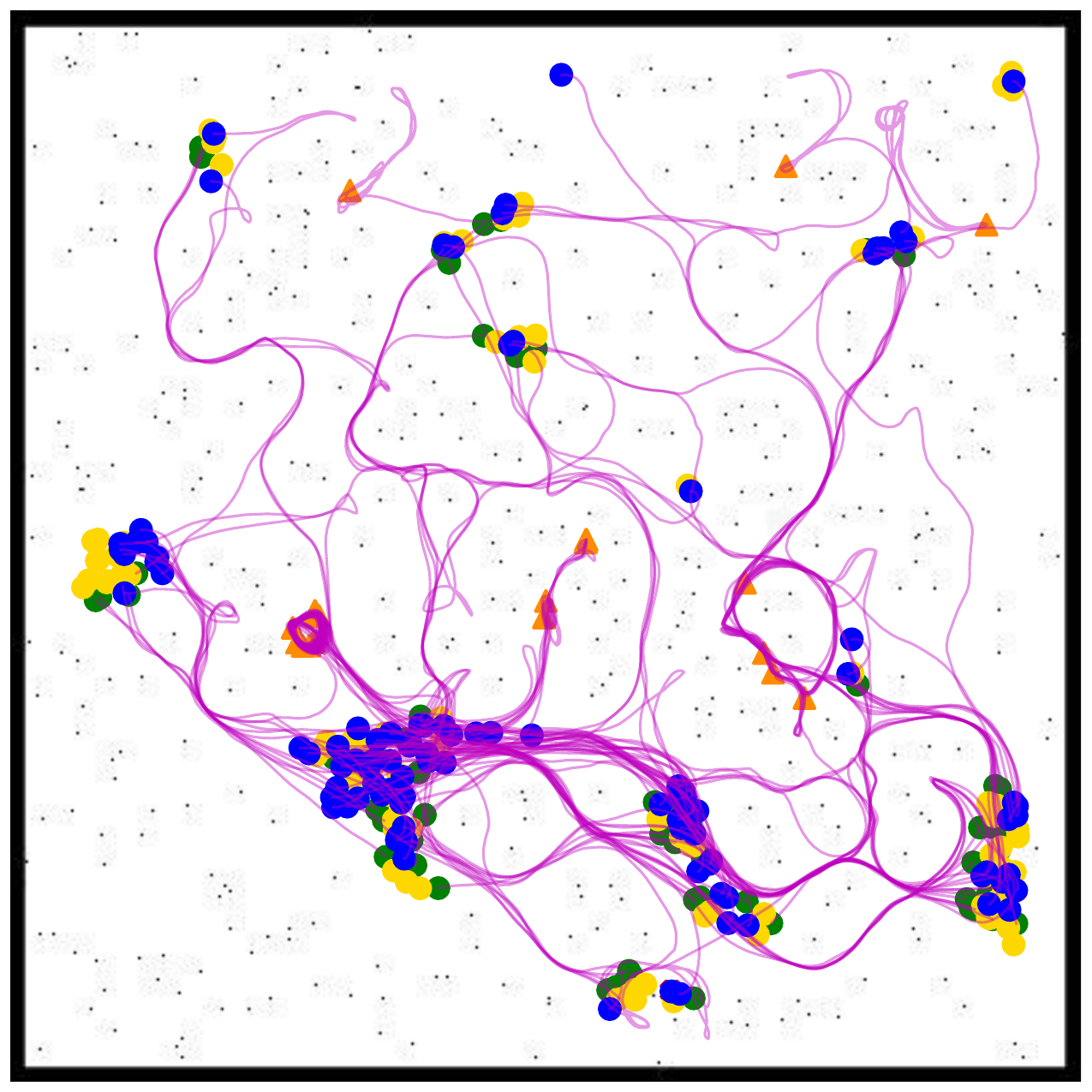}  
		\caption{SPN-v2: ${270|0.25|30|0}$}
		\label{fig:7d}
	\end{subfigure}
	\newline
	\begin{subfigure}{.24\linewidth}
		\centering
		\centerline{\includegraphics[width=.95\linewidth]{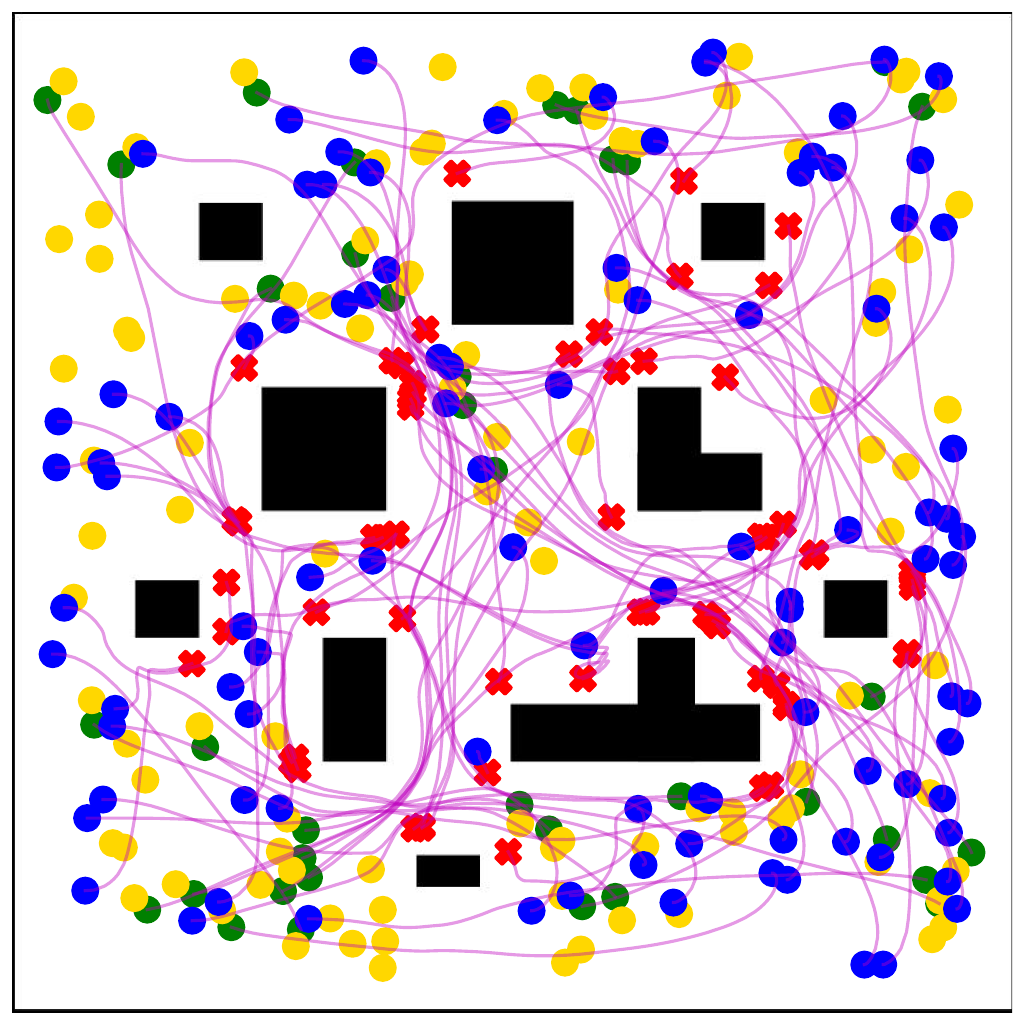}}  
		\caption{FC Net: ${180|20|10|0}$}
		\label{fig:7a}
	\end{subfigure}
	\hfill
	\begin{subfigure}{.24\linewidth}
		\centering
		\centerline{\includegraphics[width=.95\linewidth]{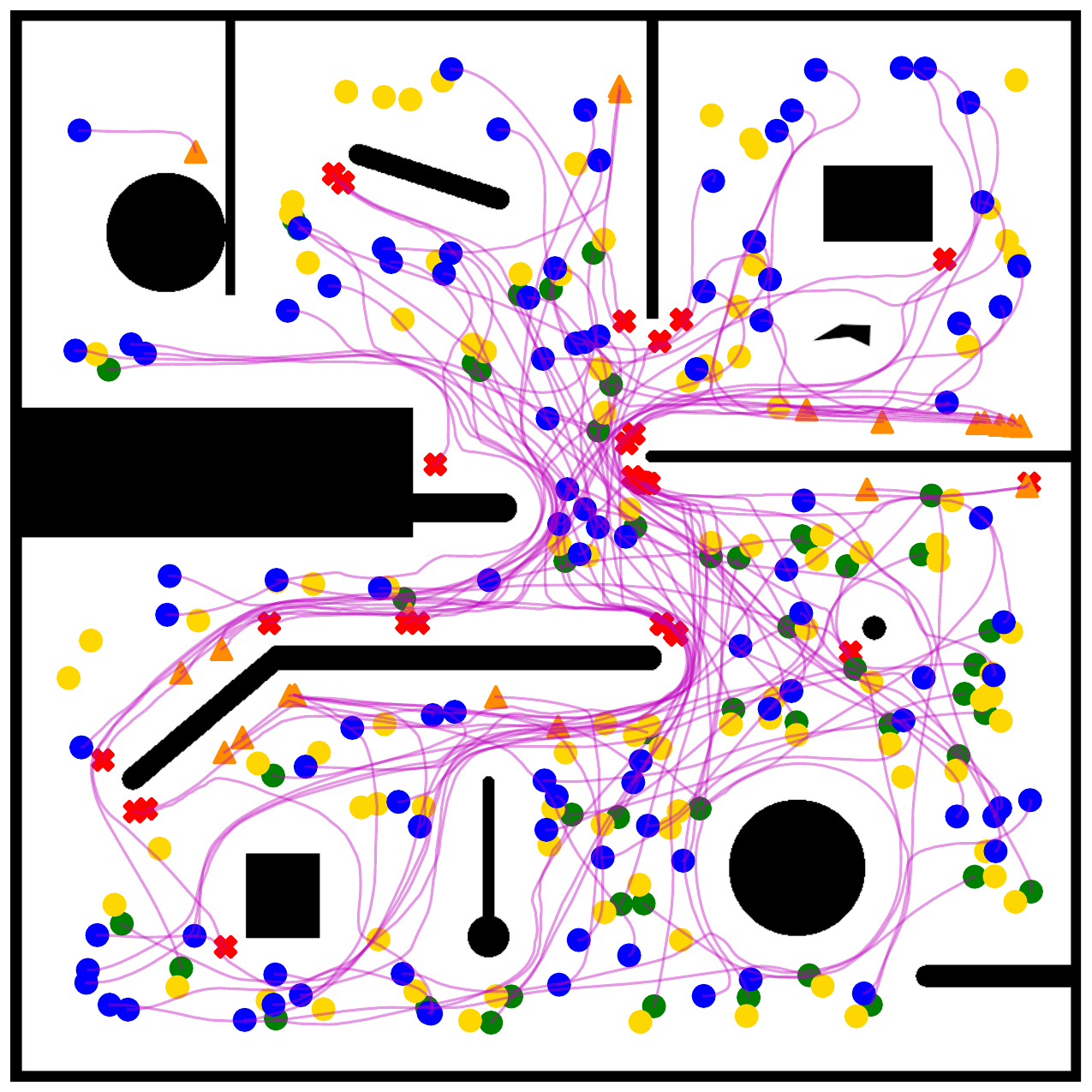}}
		\caption{FC Net: ${180|20|10|0.15}$}
		\label{fig:7b}
	\end{subfigure}
	\hfill
	\begin{subfigure}{.24\linewidth}
		\centering
		\includegraphics[width=.95\linewidth]{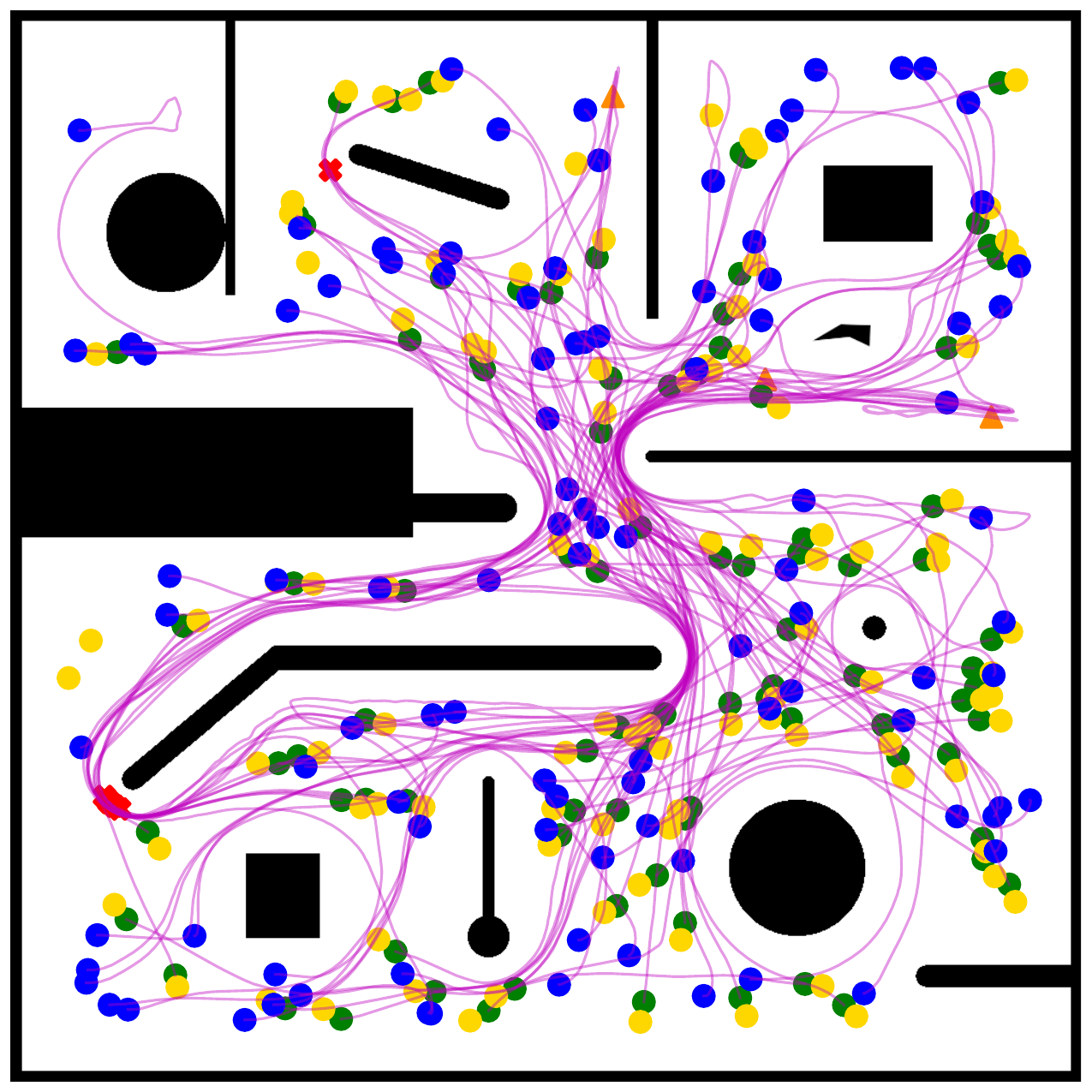}  
		\caption{FC Net: ${360|10|5|0}$}
		\label{fig:7c}
	\end{subfigure}
	\hfill
	\begin{subfigure}{.24\linewidth}
		\centering
		\includegraphics[width=.95\linewidth]{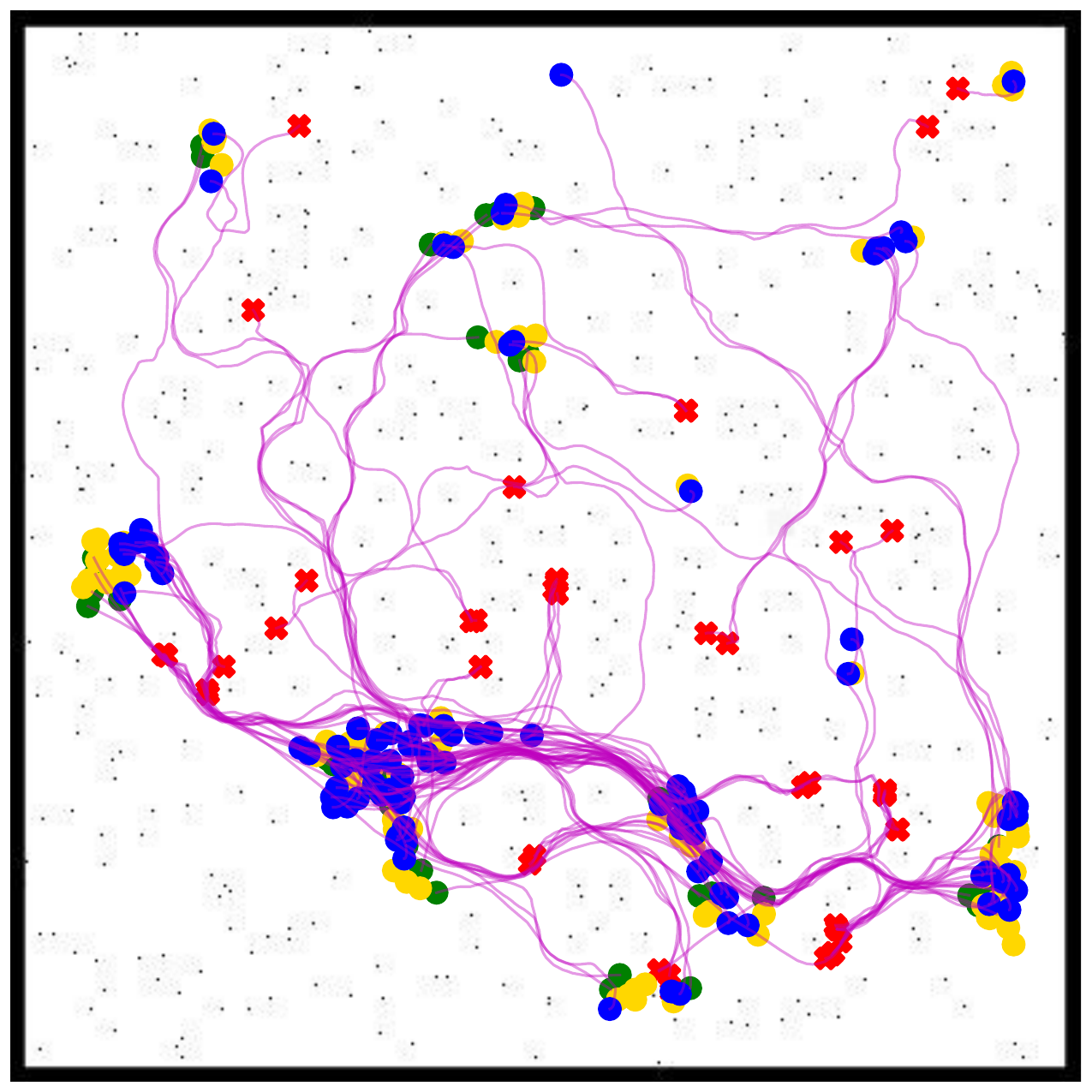}  
		\caption{FC Net: ${270|0.25|30|0}$}
		\label{fig:7d}
	\end{subfigure}
	\caption{Trajectories and terminal states visulization of robots with four different LiDAR configurations. The markers used for different terminal state are: red cross (crash), green circle (success) and orange triangle (time out). Initial and goal points are markered with blue and yellow circles, respectively.}
	\label{fig:7}
\end{figure*}
\begin{figure*}[t]
	\centering
	\includegraphics[width=0.95\textwidth]{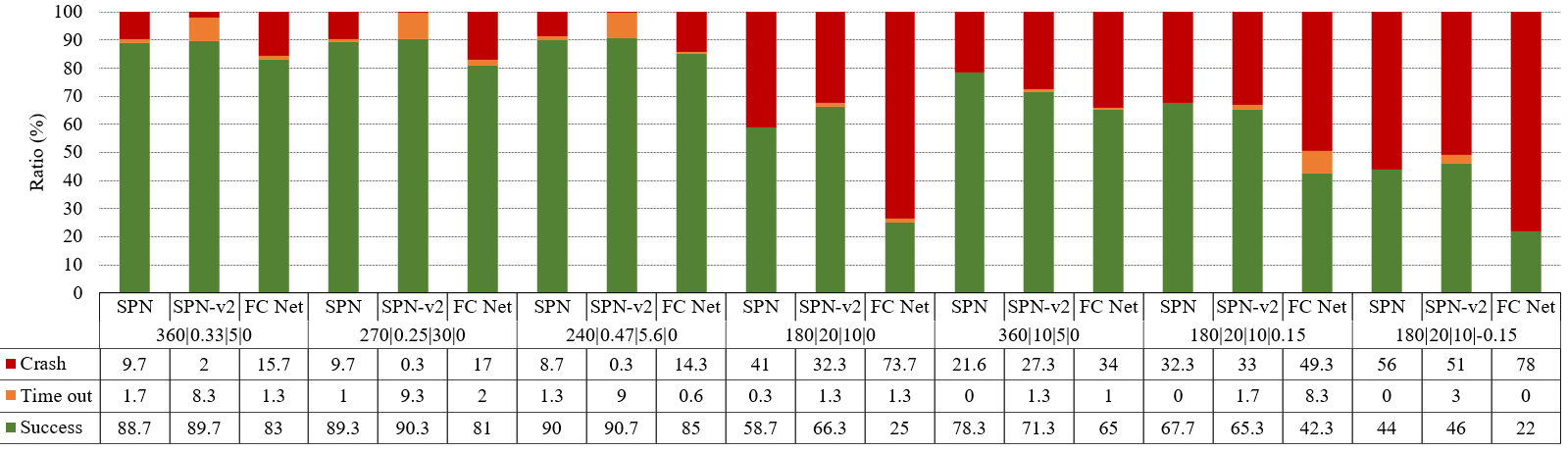}
	\caption{The averaged ratio of different ending results for each LiDAR configuration in three testing scenarios (100 runs in each scenario).}
	\label{fig:8}
\end{figure*}
\subsection{Real-world Performance Evaluation}
We test our model using a turtlebot2 robot in a real-world scenario shown in Fig. 11, in which a UTM-30LX LiDAR is used for perceiving the surroundings. As shown, the LiDAR is mounted at the top-center of the robot. Its configuration is ${``270|0.25|30|0"}$ and is different from the one used in training. During testing, at each time step, 1080 obstacle points are fed into our SPN model. The CPU in the laptop used for control is i7-7600U, and no GPU is used. As the target localization sensor, such as WIFI [22] or microphone arrays [23], is not available, same as [9], we use Gmapping [21] to pre-build a map of this scenario for robot localization. With robot and goal positions in the map, we can calculate the relative position of the goal in robot frame. It should be noted that this map is only used to calculate the relative position of the goal and is not used for any path planning by our SPN model. As shown in Fig. 11(b) and Fig. 12(a), the robot starts from the point  ``S" and is required to reach five targets ( $``\text{G}1"$ to  $``\text{G}4"$ and  $``\text{S}"$, marked with red crosses on the floor) in succession. This testing scenario contains a very small obstacle (the cylinder) and many obstacles with sharp corners. The trajectories of the robot are recorded and plotted in Fig. 12(a). As shown, the robot reaches five destinations successfully, and the chosen paths are near-optimal.
\begin{figure}[t]
	\centering
	\begin{subfigure}[h]{.32\linewidth}
		\centering
		\centerline{\includegraphics[width=0.98\linewidth]{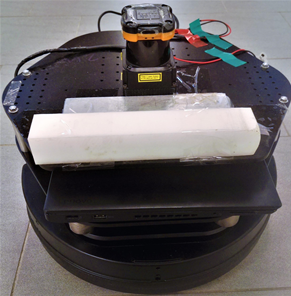}}  
		\caption{}
		\label{fig:9a}
	\end{subfigure}
	\hfill
	\begin{subfigure}[h]{.64\linewidth}
		\centering
		\centerline{\includegraphics[width=0.98\linewidth]{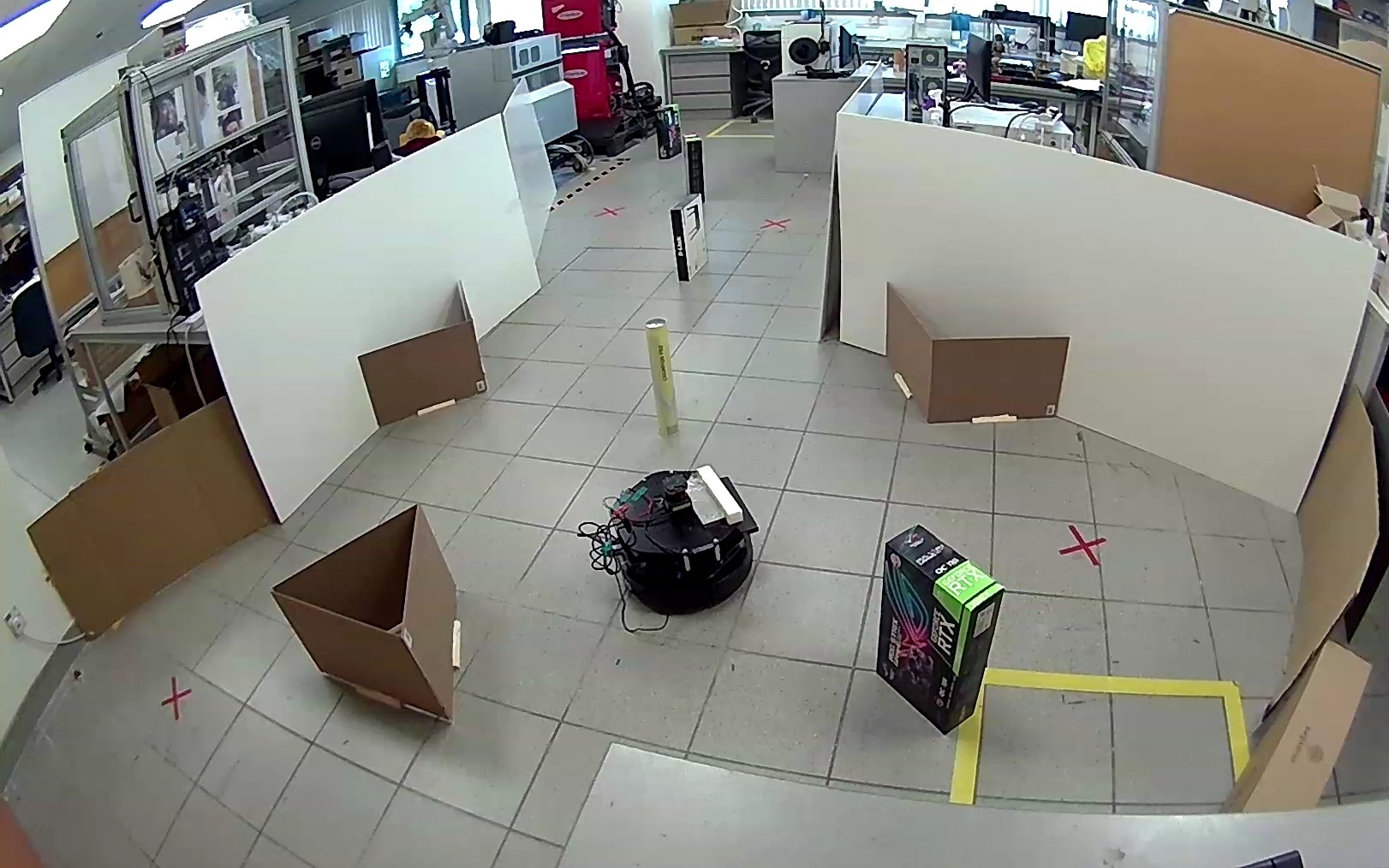}}
		\caption{}
		\label{fig:9b}
	\end{subfigure}
	\caption{Real-world testing setup. (a) A turtlebot2 and its LiDAR configurations is [$270|0.25|30|0$]. (b) Real-world testing scenarios, where the targets are marked with red crosses on the floor.}
	\label{fig:9}
\end{figure}
\begin{figure}[t]
	\centering
	\begin{subfigure}[h]{.48\linewidth}
		\centering
		\centerline{\includegraphics[width=0.98\linewidth]{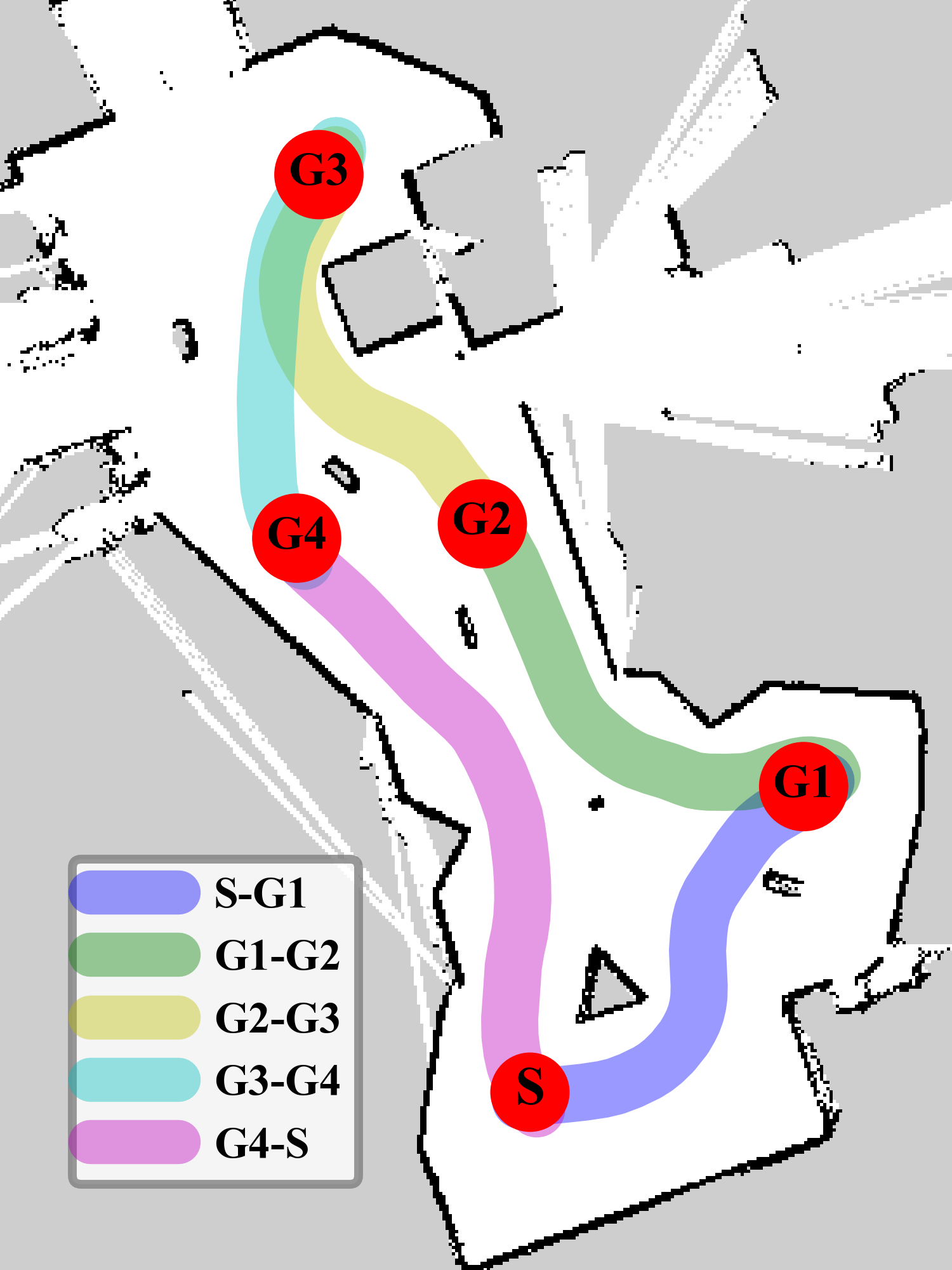}}  
		\caption{}
		\label{fig:10a}
	\end{subfigure}
	\hfill
	\begin{subfigure}[h]{.48\linewidth}
		\centering
		\centerline{\includegraphics[width=0.98\linewidth]{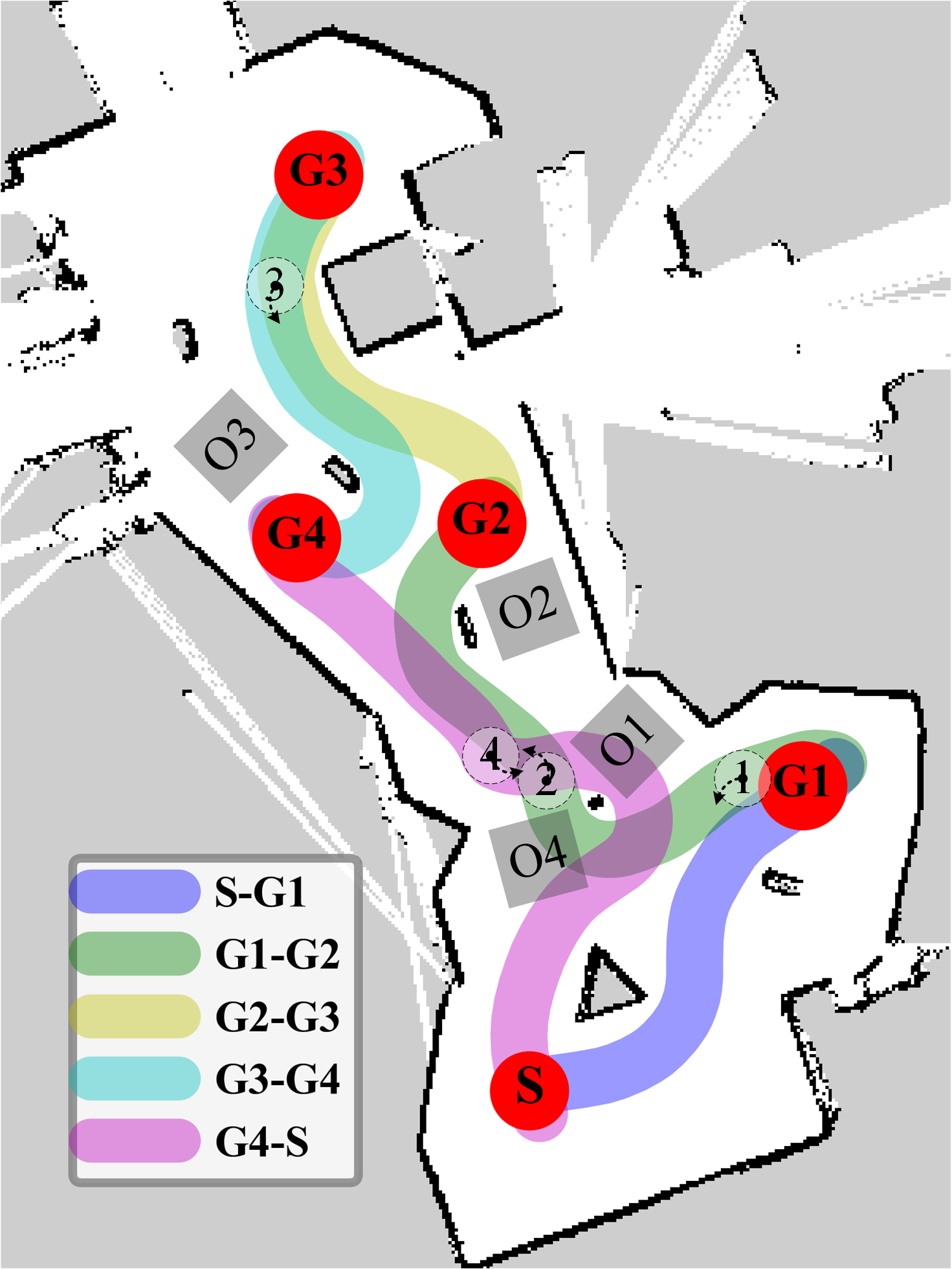}}
		\caption{}
		\label{fig:10b}
	\end{subfigure}
	\caption{Navigation trajectories of turtlebot2 robot. (a) Trajectories in the static testing scenario, (b) Trajectories in the testing scenario with sudden appearing obstacles.}
	\label{fig:10}
\end{figure}
\subsubsection{Reaction to sudden changes}
Last, we test the model's capability of addressing a suddenly appearing obstacle. As shown in Fig. 12(b) and the attached video, the researcher (labeled with  $``\text{O}i"$ in the map), serving as a dynamic obstacle, suddenly blocks the way of the robot. The robot in point $``i"$ must adjust its navigation strategy to avoid potential collisions and find another path. The trajectories and video show that our robot can well adapt to such abrupt changes and quickly switch to another path.

\section{CONCLUSIONS}

In this work, we present a learning-based approach for mapless navigation with varied LiDAR configurations. Our model selects support points from the obstacle points and can make near-optimal navigation decisions. It is purely trained in simulation and can be deployed to a real robot without any fine-tuning. Extensive simulation and real-world experiments have been conducted to evaluate our model. The experimental results demonstrate that the trained model can be used for a robot with a different LiDAR setup and can achieve high navigation performance. In the future, we plan to add a memory module in our SPN model to address the local-minimum problems in robot navigation.

\addtolength{\textheight}{-12cm}   




\section*{ACKNOWLEDGMENT}

Wei Zhang would like to thank the financial support from China Scholarship Council and National University of Singapore.


\end{document}